\documentclass[11pt]{article}

\usepackage[margin=1in]{geometry}

\usepackage{color}
\usepackage{amsfonts}
\usepackage{graphics}
\usepackage{amsmath}
\usepackage{graphicx}
\usepackage{multirow}
\usepackage{pifont,placeins,bbm}
\usepackage{enumitem}
\newtheorem{theorem}{Theorem}

\usepackage{lscape,natbib}

\title{A Huber loss-based super learner with applications \\ to healthcare expenditures}

\author
{Ziyue Wu and David Benkeser \\
Department of Biostatistics and Bioinformatics, Emory University
}
\date{}
\begin{document}

\maketitle

\begin{abstract}
Complex distributions of the healthcare expenditure pose challenges to statistical modeling via a single model. Super learning, an ensemble method that combines a range of candidate models, is a promising alternative for cost estimation and has shown benefits over a single model. However, standard approaches to super learning may have poor performance in settings where extreme values are present, such as healthcare expenditure data. We propose a super learner based on the Huber loss, a ``robust'' loss function that combines squared error loss with absolute loss to down-weight the influence of outliers. We derive oracle inequalities that establish bounds on the finite-sample and asymptotic performance of the method. We show that the proposed method can be used both directly to optimize Huber risk, as well as in finite-sample settings where optimizing mean squared error is the ultimate goal. For this latter scenario, we provide two methods for performing a grid search for values of the robustification parameter indexing the Huber loss. Simulations and real data analysis demonstrate appreciable finite-sample gains in cost prediction and causal effect estimation using our proposed method.\vspace{1em}

Keywords: healthcare expenditure, Huber loss, outliers, robust regression, super learning
\end{abstract}

\maketitle

\section{Introduction}

The study of healthcare expenditures is a key component of health services research. A common goal in studies of expenditures is to model the conditional mean cost Y given demographic, clinical, and other information. Such models can be used both to predict future healthcare expenditures, as well as to estimate effects of policy interventions on mean expenditures \citep{basu2011estimating, smith2016estimating}. However, statistical modeling of healthcare expenditures is often challenging due to extreme expenditures in a small number of individuals. In 2009, for example, 22\% of total health care expenditures in the United States were allocated to just 1\% of the U.S. population, and almost 50\% of health care spending was devoted to 5\% of the population \citep{cohen2012statistical}. Outliers in costs can disproportionately influence prediction models, biasing predictions towards outliers and limiting the model accuracy \citep{weichle2013impact}.

Many methods have been proposed to deal with these challenges \citep{jones2010models}. A common approach involves regression of log-transformed cost using ordinary least squares. Duan’s smearing estimator \citep{duan1983smearing} can be used to map estimates from the log-scale back to the original scale. Another approach is the use of non-Gaussian GLMs \citep{manning2001estimating}, which provide greater flexibility by allowing the outcomes to follow a non-normal distribution. Alternatively, robust regression can be used. One popular robust approach is Huber regressions, which rely on minimizing a loss function (Huber loss) that assigns outliers less weight \citep{Huber1973robust, huber1992robust}. However, in Huber regression the robustness against outliers, controlled by a robustification parameter $\lambda$, is gained at the price of accuracy loss when $\lambda$ is unnecessarily high. Therefore, accurate estimation is possible only if an appropriate $\lambda$ is chosen. Little guidance is available on how to choose an appropriate robustification parameter in practice.

In addition to these more traditional regression-based approaches described above, there has recently been increased interest in machine learning techniques. These methods use flexible approaches to learn about relationships in data, thereby providing, in many cases, more accurate cost predictions \citep{gareth2013introduction, rose2016machine, morid2017supervised, shrestha2018mental, zink2020fair}.


Because there is no universally accepted ``best approach'' to dealing with extreme expenditure data, it is desirable to have a robust, empirical method for selecting amongst estimators. In this work, we propose an ensemble method that combines a range of candidate models via super learning (also known as model stacking) that is tailored towards data with skewed outcomes like healthcare expenditures \citep{wolpert1992stacked,  breiman1996stacked, van2007super}. The existing super learner literature provides a general framework for using cross-validation to learn the optimal combination of a user-specified collection of candidate methods. This collection could include parametric and machine learning-based algorithms. Weights for each algorithm are chosen via a step of ``meta-learning'', which entails minimizing a user-specified cross-validated risk criteria, such as mean squared error, over some class of weights. Super learner has shown benefits over a single method in various healthcare studies \citep{rose2013mortality, kessler2014well, pirracchio2015mortality} including prediction of expenditures in the context of plan payment risk adjustment \citep{rose2016machine}. A two-stage super learner has also been developed for data that exhibit zero-inflation \citep{wu2021two}.

The contribution of this work is to propose a new risk criterion that can be used in the meta-learning step of super learner in lieu of squared error loss. In particular, we propose a super learner based on the Huber loss, which combines the typical squared error loss with absolute loss to down-weight the influence of outliers. This loss is indexed by a robustification parameter, which determines where the switch from squared error to absolute error occurs. We prove that, under regularity conditions, the Huber loss-based super learner enjoys an oracle property: the performance of the super learner in terms of Huber risk is essentially equivalent with that of the oracle selector. Thus, the proposed method is an optimal means of estimator selection and/or model stacking in settings where Huber risk is directly of interest. Moreover, we also argue that the Huber risk could be used as a ``working loss function,'' in settings where mean squared error is ultimately of interest. An example of the latter setting is one where the super learner would be used as an intermediate step in drawing inference about the average effect of a policy intervention on expenditures. To this end, we propose two methods for performing a grid search for values of the robustification parameter. We demonstrate the benefits of the proposed method relative to super learners based directly on squared error loss using simulated data, as well as two real data sets of healthcare expenditures, the 2016-2017 Medical Expenditure Panel Survey (MEPS) \citep{cohen2003design, cohen2009medical}, and Back pain Outcomes using Longitudinal Data (BOLD) project \citep{jarvik2012study}. 

\section{Methods}

\subsection{Super learning}

Suppose we have a dataset of identically distributed observations $O_i = (Y_i, X_i) \sim P, i=1, \dots ,n$ where $Y$ is the outcome of interest and $X$ is a set of covariates that will be used to predict the outcome. Without loss of generality, we assume $\mathcal{Y}$ has support on $\mathbb{R}^+$ and denote by $\mathcal{X}$ the support of $X$. For simplicity, we will assume the observations are independent, though super learning procedures have been developed for correlated data as well \citep{benkeser2018online}. Suppose we have access to a candidate prediction function $\hat{Q}: \mathcal{X} \rightarrow \mathbb{R}$. In a given learning context, we can define a scientifically relevant \textit{risk} criteria, $R(\hat{Q})$, that describes the global performance of $\hat{Q}$ for predicting the outcome $Y$ based on covariates $X$. For example, mean squared error $R(\hat{Q}) = \int \{y - \hat{Q}(x) \}^2 dP(o)$ is often selected as a risk criteria. Many risk criteria that are commonly used in practice can be expressed as the average value of a \textit{loss function}, $L(\hat{Q})$, a function that measures the discrepancy for a single datum $o = (x,y)$ between the prediction $\hat{Q}(x)$ made and the actual outcome $y$. For example, mean squared error is the average value of the squared error loss function $L(\hat{Q})(o) = \{y - \hat{Q}(x)\}^2$. Given a risk criterion, the goal of statistical learning can often be stated as approximating the optimal prediction function, say $Q_0$, defined as the minimizer of $R(\hat{Q})$ over all choices of prediction function $\hat{Q}$.


In super learning, we propose many \textit{algorithms} for learning $Q_0$ and refer to this pre-specified collection of algorithms as a \textit{library}. Here, ``algorithm'' is used in a general sense as any means of mapping a given data set into a prediction function. Examples of such algorithms include (i) fitting ordinary least squares regression and returning the linear predictor; (ii) performing variable screening based on a univariate significance threshold, then applying ordinary least squares regression; (iii) training a random forest, where tuning parameters are selected via cross-validation. It is recommended that the library should be informed by subject-area expertise, but can utilize data-driven, machine learning approaches as well \citep{benkeser2020rejoinder}.

Suppose we have $K$ potential learning algorithms. Super learner is implemented as follows.\begin{enumerate}
    \item Train each algorithm in the library on the entire dataset yielding prediction functions $\hat{Q}_k, 
\ k=1, \dots ,K$.
\item Randomly split the data into $V$ mutually exclusive and exhaustive blocks of approximately equal size. For $v=1, \dots ,V$, define the $v$-th block as the $v$-th \textit{validation sample}, and the remaining $V-1$ blocks the $v$-th \textit{training sample}. 
\item For $v=1, \dots ,V$, train each of the $K$ algorithms using the $v$-th training sample yielding prediction functions $\hat{Q}_{k, v}$.
\item Given the $K$ trained algorithms from the $v$-th validation sample, propose an ensemble and define a parameter space $\mathcal{A}$ for the weight vector $\alpha$. For example, we could propose an ensemble based on a convex combination of the algorithms:
$$
\hat{Q}_{\alpha, v}^{\text{SL}}=\sum_{k=1}^{K} \alpha_{k}\hat{Q}_{k,v},\; \mathcal{A} =\{\alpha : \alpha_{k}\geq 0 \ \forall \ k, \ \sum_{k=1}^{K}\alpha_{k} = 1\} \ .
$$
\item Compute $\hat{\alpha} = \mbox{arg}\mbox{min}_{\alpha \in \mathcal{A}} \sum_{v=1}^V \hat{R}(\hat{Q}_{\alpha, v}^{\text{SL}}),$ the weights that minimize an estimate of cross-validated risk. For example, if $R$ is mean squared error, then denoting by $\mathcal{I}_v$ the indices of observations in the $v$-th validation sample, we would use $\hat{R}(\hat{Q}^{\text{SL}}_{\alpha, v}) = \frac{1}{|\mathcal{I}_v|}\sum_{i \in \mathcal{I}_v} \{ Y_i - \hat{Q}^{\text{SL}}_{\alpha, v}(X_i) \}^2.$
\item The super learner is defined as $\hat{Q}_{\hat{\alpha}}^{\text{SL}} = \sum_{k=1}^{K}\hat{\alpha}_k \hat{Q}_k$.
\end{enumerate}

A particular instantiation of the super learner is defined by the choice of library, cross-validation scheme, risk criteria, and ensemble. One particular choice is to build the ensemble from a set of weight vectors that each assigns a weight of 1 to a particular algorithm and 0 weight to all others. This choice is known as the cross-validation selector or discrete learner since it represents the single algorithm with the lowest cross-validated risk.

Typically, when the outcome $Y$ is real-valued, super learner is built using mean squared error as the risk criteria, as evidenced by, for example, this choice being the default in common super learner software implementations. We will thus refer to this approach as the \textit{standard} super learner. Oracle inequalities have been established \citep{van2003unified, van2006cross, van2007super} to show that in large samples the super learner should predict costs essentially as well as or better than the best-performing single algorithm amongst the $K$ considered in terms of mean squared error. However, in finite samples, outliers may have an adverse impact on the selection of weights in the standard super learner, leading to a degraded overall performance of the ensemble. In the next section, we propose a robust risk criteria for improving the super learner in the context of data with outliers.

\section{Huber loss-based super learner}

\subsection{Huber loss function and super learner}

We propose to develop a super learner based on the Huber risk function, which is based on a blended quadratic and linear loss defined for a particular $\hat{Q}$ and data realization $o = (x,y)$ as 
\begin{equation}
L_{\lambda}(\hat{Q})(o)=\left\{\begin{matrix}
\frac{1}{2}\left\{y-Q(x)\right\}^2 & |y-Q(x)| \leq \lambda  \\
\lambda(|y-Q(x)| - \frac{1}{2}\lambda) & |y-Q(x)| > \lambda
\end{matrix}\right.\ , \ \lambda \in \mathbb{R}^+
\end{equation}

where $\lambda>0$ is referred to as the robustification parameter that can be used to regulate the amount of influence of outliers. The loss function $L_{\lambda}$ becomes more similar to squared error loss for larger value of $\lambda$ while it becomes more similar to absolute loss for smaller values of $\lambda$. Huber risk is then defined as $R_{\lambda}(\hat{Q})=\int L_{\lambda}(\hat{Q})(o) dP(o)$. 

We propose to base a super learner on the Huber loss. That is, given a particular $\lambda$, we propose to find weights $\hat{\alpha}$ that minimize the cross-validated estimate of the Huber risk, defined as:
\begin{equation}
\hat{\alpha} = \mbox{arg} \ \mbox{min}_{\alpha \in \mathcal{A}} \sum_{v=1}^V \hat{R}_{\lambda}(\hat{Q}_{\alpha, v}^\text{SL})
\end{equation}
where $\hat{R}_{\lambda}(\hat{Q}_{\alpha, v}^\text{SL}) = \frac{1}{|\mathcal{I}_v|} \sum_{i \in \mathcal{I}_v} L_{\lambda}(\hat{Q}_{\alpha, v}^{\text{SL}})(O_i).$

\subsection{Oracle inequality}


Without loss of generality, we consider $V$-fold cross-validation and denote by $\hat{Q}_{k,v},\ k=1,\dots,K, \ v=1,\dots,V$ the realization of individual algorithm $k$ fitted on the $v$-th training sample and $\hat{Q}_{\alpha,v}^\text{SL}$ as the super learner fitted on the $v$-th training sample. Consider an oracle super learner $\hat{Q}^\text{SL}_{\tilde{\alpha}}$ with oracle weights $\tilde{\alpha}$ that minimize the true Huber risk given $V$ training-sample-specific fits:
$$\tilde{\alpha}=\mbox{arg min}_{\alpha \in \mathcal{A}} \sum_{v=1}^V \int L_{\lambda}(\hat{Q}_{\alpha, v}^\text{SL})(o) dP(o) = \mbox{arg min}_{\alpha\in \mathcal{A}} \sum_{v=1}^V R_{\lambda}(\hat{Q}_{\alpha, v}^\text{SL}) \ .$$
The oracle weights represent the best weights that could be selected at a population-level, given the training samples seen. Such oracle weights $\tilde{\alpha}$ do not need to be unique. Note that the oracle weights depend on the unknown true distribution $P$, hence are not generally available in practice. We now present a theorem that compares the risk of the Huber loss-based super learner $\hat{Q}^\text{SL}_{\hat{\alpha}}$ to that of the oracle super learner $\hat{Q}^\text{SL}_{\tilde{\alpha}}$ built with oracle weights $\tilde{\alpha}$.

\begin{theorem} Consider the Huber loss function $L_{\lambda}(\hat{Q})$ defined for a given $\lambda \in \mathbb{R}^+$, and a collection of super learners in $\mathcal{Q}=\{\hat{Q}^\text{SL}_{\alpha}: \alpha \in \mathcal{A}_n\}$, where $\hat{Q}^\text{SL}_{\alpha} = \sum_{k=1}^K \alpha_k\hat{Q}_k$. Assume that \begin{enumerate}
    \item[(A1)] The parameter space $\mathcal{A}$ is bounded and $\mathcal{A}_n$ is a finite subset of $\mathcal{A}$;
    \item[(A2)] the number of grid points in $\mathcal{A}_n$ (candidate super learner weights) grows at most at polynomial rate in $n$: $K(n) = |\mathcal{A}_n| \leq n^q$ for some $q < \infty$; and
    \item[(A3)] both individual algorithms $\hat{Q}_k,\ k=1,\dots,K$ and outcome $Y$ are bounded: $|y|\leq C_0$ for all $y \in \mathcal{Y}$ and $\mbox{sup}_{k}\mbox{sup}_{x \in \mathcal{X}}|\hat{Q}_k(x)| \leq C_0$ for some $C_0 < \infty$.
\end{enumerate}
Under these conditions, we have \begin{equation*}
\sum_{v=1}^V R_{\lambda}(\hat{Q}_{\hat{\alpha}, v}^\text{SL}) \leq \sum_{v=1}^V R_{\lambda}(\hat{Q}_{\tilde{\alpha}, v}^\text{SL}) + O\left(\frac{\log(n)}{n^{1/2}}\right) \ .
\end{equation*}
where $R_{\lambda}(\hat{Q}_{\hat{\alpha}, v}^\text{SL})$ and $R_{\lambda}(\hat{Q}_{\tilde{\alpha}, v}^\text{SL})$ are the Huber risks averaged over $V$ training-sample-specific fits for the Huber loss-based super learner and the oracle super learner respectively.
\end{theorem}

Assumption (A1) stipulates that the parameter space $\mathcal{A}$ for the weights is bounded, which places only a minor restriction on the types of meta-learning that are supported under this theorem. Assumption (A2) requires a control on the cardinality of candidate super learners and is a standard assumption in the super learning literature. If one considers the cross-validation selector, then $\mathcal{A}$ is a collection of $K$ weight vectors where the $k$-th vector has a 1 in the $k$-th element and zeros elsewhere. In this case (A2) requires that the number of candidate algorithms in the super learner can grow no faster than a polynomial rate. If on the other hand, true ensemble models are considered, then (A2) would require a considerably slower growth in the number of models, since linear growth in $K$ implies potentially exponential growth in $\mathcal{A}_n$. Assumption (A3) requires bounded outcomes and individual algorithms, which is a highly plausible assumption in healthcare expenditure applications. Overall, these assumptions appear reasonable in practical applications and are standard assumptions in the super learning literature.

The result of the theorem is that under the stated conditions, the Huber loss-based super learner $\hat{Q}_{\hat{\alpha}}^\text{SL}$ asymptotically converges to the oracle super learner $\hat{Q}_{\tilde{\alpha}}^\text{SL}$ in terms of averaged risk as the sample size grows to infinity. The theorem extends previous results (e.g., \citet{van2007super}) pertaining to super learners constructed based on squared error loss to Huber loss. Details of the notations and proof are provided in the Appendix A.1 of the supplementary material. In brief, the proof proceeds by showing that the Huber loss function is uniformly bounded and creating a Bernstein pairs for the Huber loss function. A tail inequality for Bernstein pairs is then used to derive the result.

\subsection{Selecting robustification parameter}

The oracle inequality above establishes the optimality of the Huber loss-based super learner in terms of \emph{Huber risk}. In some settings, optimizing Huber risk may be directly of interest if a suitable, scientifically relevant value of robustification parameter $\lambda$ can be selected a-priori. In other settings, our purpose may be only to use the Huber loss as a means of increasing the robustness of the super learner ensemble to outlying values, while ultimately our interest is in an alternative risk criteria, such as mean squared error. For example, suppose we are interested in estimating the average causal effect of a non-randomized binary policy decision $A$ on the average value of $Y$. Several common estimation frameworks rely on estimation of $Q(A,X) = E(Y \mid A, X)$ as an intermediate step (e.g., augmented inverse probability weighting \citep{robins1994estimation} and targeted minimum loss estimation \citep{van2006targeted}). Because the conditional mean $Q$ is the minimizer of mean squared error, we may want to use a super learner based on squared-error loss to estimate $E(Y \mid A, X)$ thereby harmonizing the choice of risk criteria with the parameter targeted by the procedure. However, if the conditional distribution of $Y$ contains heavy tails, in finite-samples, we may wish to use the Huber loss as a \textit{working} loss function in order to increase the robustness of our results.

In such settings, we suggest proposing a sequence of $\lambda$ values for the Huber loss robustification parameter, building super learner ensembles based on these Huber losses, and finally using a data-driven procedure to select from among the various ensemble weights the weights which minimize a cross-validated mean squared error criterion. We propose two such procedures, based on \emph{partial} cross-validation (partial-CV) and \emph{nested} cross-validation (nested-CV).

The procedure based on partial cross-validation proceeds as follows. We first propose a collection of candidate robustification parameters $\lambda_1, \dots ,\lambda_J$. For each $\lambda_j, j = 1,\dots,J$, we compute Huber loss-based super learner weights $\alpha_j, j = 1,\dots,J$ as above $\hat{\alpha}_j = \mbox{arg min}_{\alpha} \sum_{v=1}^V \hat{R}_{\lambda_j}(\hat{Q}_{\alpha, v}^{\text{SL}})\ , \ j = 1,\dots,J.$
The Huber loss-based super learner using weights $\alpha_j, \ j = 1, \dots, J$ is then defined as $\hat{Q}_{\alpha_j, v}^{\text{SL}} = \sum_{k=1}^K \hat{\alpha}_{j, k}\hat{Q}_{k, v} \ , \;j = 1,\dots,J$. Next, we select the  robustification parameter that minimizes a partially cross-validated mean squared error, $\hat{j} = \mbox{arg min}_{j = 1, \dots, J} \sum_{v=1}^V \sum_{i \in \mathcal{I}_v} \{Y_i - \hat{Q}_{\hat{\alpha}_j, v}^{\text{SL}}(X_i)\}^2.$
The final Huber loss-based super learner is then $\hat{Q}_{\hat{\alpha}_{\hat{j}}}^{\text{SL}} = \sum_{k=1}^{K}\hat{\alpha}_{\hat{j}, k} \hat{Q}_k$.

The computing time for this procedure is identical to a standard super learner implementation, with the time largely driven by the number of estimators fitted during the cross-validation procedure. However, this procedure is not a ``proper'' cross-validation procedure, since the full data are used both in the calculation of super learner weights and selection of robustification parameter. Thus, we may run the risk of over-fitting, particularly in settings where the set of candidate learners is large.

To reduce over-fitting, we can use proper, nested cross-validation for calculation of robustification parameter. Given a collection of candidate robustification parameters $\lambda_1, \dots ,\lambda_J$, we first define the \textit{outer} cross-validation folds by dividing the entire data set into $V$ mutually exclusive training and validation samples. For each training sample, we further conduct a $D$-fold \textit{inner} cross-validation to compute each of the Huber loss-based super learner weights $\alpha_{j,v}$ corresponding to each $\lambda_j, j = 1,\dots,J$, denoted as $\hat{\alpha}_{j,v} = \mbox{arg min}_{\alpha\in \mathcal{A}} \sum_{d=1}^D \hat{R}_{\lambda_j}(\hat{Q}_{\alpha, d}^{\text{SL}})$. Thus, we train $V \times J$ total Huber loss-based super learners, with the super learner trained in the $v$-th training sample with $\lambda_j$ as the value of the robustification parameter defined as $\hat{Q}_{\hat{\alpha}_{j,v}}^{\text{SL}} = \sum_{k=1}^K \hat{\alpha}_{j,v,k}\hat{Q}_{k, v} \ , \;j = 1,\dots,J \ , \;v = 1,\dots,V.$ With a slight abuse of notation, we define the proper cross validation selector of the robustification parameter as $\hat{j} = \mbox{arg min}_{j = 1, \dots, J} \sum_{v=1}^V \sum_{i \in \mathcal{I}_v} \{Y_i - \hat{Q}_{\hat{\alpha}_{j,v}}^{\text{SL}}(X_i)\}^2.$ Given the selected robustification parameter $\lambda_{\hat{j}}$, we compute the super learner weights $\hat{\alpha}_{\lambda_{\hat{j}}} = \mbox{arg min}_{\alpha\in \mathcal{A}} \sum_{v=1}^V \hat{R}_{\lambda_{\hat{j}}}(\hat{Q}_{\alpha, v}^{\text{SL}}).$ 

A drawback of the nested cross-validation procedure is the increase in the computing time and number of estimators fitted in cross-validation. However, we anticipate that this procedure may lead to performance improvements over the partial cross-validation strategy above.

\section{Monte Carlo Simulation}

We evaluated the performance of the proposed method using Monte Carlo simulations. We were interested in assessing performance of the method in two tasks that are common in biomedical cost research: cost prediction and estimation of effects of interventions on costs. In both cases, we focus on the use case of our method where the Huber loss is used as a \emph{working} loss function, as we anticipate these situations may arise more readily in practice. In both simulation studies, we used the same data generating process for $X$. The vector $X$ consisted of ten components, distributed as $X_1 \sim \mbox{Bernoulli}(0.5)$, $X_2 \sim \mbox{Uniform}(0,1)$, $X_3 \sim \mbox{Normal}(0,1)$, $X_4 \sim \mbox{Gamma}(1,1)$, $X_5 \sim \mbox{Poisson}(1)$, $X_6 \sim \mbox{Bernoulli}(0.2)$, $X_7 \sim \mbox{Uniform}(-1,1)$, $X_8 \sim \mbox{Normal}(0,3)$, $X_9 \sim \mbox{Gamma}(0.5,1)$, $X_{10} \sim \mbox{Poisson}(2)$. In both simulations, the variables $X_1, \dots , X_5$ impacted the distribution of costs, while the others were noise. The other aspects of the data generating process differed by simulation setting and are described below.

\subsection{Huber super learner for cost prediction}

The goal of this simulation was to compare the proposed Huber loss-based super learners with existing implementations across a variety of data generating settings. In particular, for a fixed zero-percentage ($\sim$35\%), we were interested in comparing performance of the methods under varying sample sizes and varying skewness in the outcome data. While there are various ways of quantifying skewness, in this study we quantify skewness in terms of the fraction of observations that qualify as ``outliers'', that is, observations that are larger than the upper quartile of the data plus 1.5 times the interquartile range. We considered settings with low ($\sim$3.5\%), medium ($\sim$10\%), and high ($\sim$20\%) outlier proportions and compared performance of methods at sample sizes of 250, 500, 1000, and 2000. Thus, we consider a total of twelve scenarios. For each scenario, we analyzed 1000 simulated data sets.

To generate simulated costs, we used a two-stage simulation procedure to allow for point mass at zero, a common aspect of cost distributions encountered in practice. In the first stage, we drew a random variable $Z$ from a Bernoulli distribution with the probability of zero determined by a logistic model (Appendix A.2 of the supplementary material). If $Z=0$, we set costs to zero. If $Z=1$, we drew the observed costs from a gamma distributions with scale parameter fixed at 1.5 and shape parameter that was a function of $X$, where the functional form of the shape parameter varied across the outlier settings (Appendix A.2 of the supplementary material). For simplicity in determining the grid of $\lambda$ used by our method in the simulation, a scaling procedure was applied  to ensure that all the simulated cost data are between 0 and 1$e$6.

We considered four candidate algorithms for the super learner: ordinary least squares (OLS), lasso \citep{tibshirani1996regression}, Support Vector Machine (SVM) \citep{cortes1995support} and Random Forest \citep{breiman2001random}. These four algorithms were used to define six super learners: standard discrete super learner (based on optimizing MSE), standard super learner ensemble (based on optimizing MSE), partial-CV Huber loss-based discrete super learner, nested-CV Huber loss-based discrete super learner, partial-CV Huber loss-based super learner ensemble, and nested-CV Huber loss-based super learner ensemble. The Huber loss-based learners used 29 candidate values of $\lambda$ ranging from 0.1 to 1,000,000 for $\lambda$ selection. All super learners were based on ten-fold cross-validation. 

Cost prediction performances were evaluated on an independent test set of size $n_{test} = 10000$. Let $y_i$ be the true outcome values on the test set, $\hat{y}_i$ be the predicted outcome values from certain algorithms on the test set and $\bar{y}_i$ be the mean of $y_i$, we use the following metrics for evaluation: mean squared error, $
\text{MSE} = \frac{1}{n_{test}}\sum_{i=1}^{n_{test}}(y_i - \hat{y}_i)^2
$; $\text{relative MSE} = \frac{\text{MSE (Huber loss-based learner})}{\text{MSE (standard learner)}}$; mean absolute error, $\text{MAE} = \frac{1}{n_{test}}\sum_{i=1}^{n_{test}}|y_i - \hat{y}_i|$; coefficient of determination, $R^2 = 1 - \frac{\sum_{i=1}^{n_{test}}(y_i - \hat{y}_i)^2}{\sum_{i=1}^{n_{test}}(y_i - \bar{y}_i)^2}$; relative efficiency, $\text{RE} = \frac{R^2 (\text{Huber loss-based learner})}{R^2 (\text{standard learner})}$.

Pooling across the twelve settings, the use of the Huber loss in the construction of the super learner tended to result in superior performance over use of MSE (Table 1). Focusing on the super learner, the MSE improvement over the standard super learner ranges from 2.2\% (partial-CV Huber loss-based super learner) to 4.4\% (nested-CV Huber loss-based super learner). For $\lambda$ selection, nested-CV is preferred over partial-CV with further improvement in cost prediction. Improvements in $R^2$ for nested-CV over partial-CV were about 5.8\% for Huber loss-based discrete learner and 4.5\% for Huber loss-based super learner. Although prediction performance elevated in large sample size and low skewness, the superior performance of the Huber loss-based learner over the standard learners remained the same under different level of sample size and skewness (Appendix Tables 4 and 5 of the supplementary material).

Figure 1 displays the MSE ratio (log-scale on Y axis) of one-stage Huber loss-based learners (discrete learner \& super learner) over the one-stage standard learners (discrete learner \& super learner) under different skewness and sample size. MSE ratio smaller than 1 indicates better predictions compared to the standard learners and lower MSE ratio means further improvement over the standard learners. Similarly, Huber loss-based learners, especially the nested-CV Huber loss-based learners, had better risks compared to the standard learners with majority of MSE ratio below 1. Furthermore, across all settings, the improvements of the Huber loss-based learners over the standard learners are larger under higher skewness and smaller sample size, though cost predictions tend to be more unstable in these cases as shown by the more scattering points under high skewness and small sample size in Figure 1.

An additional simulation is included in the Appendix A.3 of the supplementary material, where we evaluate and compare the six super learners based on a more extensive library and a two-stage model building approach. The conclusions of the simulation are the same as for the results presented here (Appendix Table 6 and Appendix Figure 1 of the supplementary material).

\subsection{Estimation of the average treatment effect}

We conducted another simulation to evaluate the Huber loss-based super learner for the purpose of estimation of the average treatment effect (ATE) in the context of a randomized controlled trial. For data generation, the same covariate vector $X$ was used. The variable $X_1 \sim Bernoulli(0.5)$ was considered a randomized intervention. Because treatment is randomized, inference on the ATE may be facilitated through a comparison of the sample average cost compared those with $X_1 = 1$ and $X_1 = 0$. However, precision can be increased in this context by further adjusting this comparison for covariates \citep{senn1989covariate, robinson1991some, tsiatis2008covariate}. Thus, we considered targeted minimum loss estimators (TMLEs) that adjust for variables $X_2, \dots, X_5$ via super learning of the conditional mean of costs given $X_1, \dots, X_5$. In line with previous research, we hypothesized that TMLEs would exhibit smaller variance than the difference in sample means. We further hypothesized that the usage of Huber loss-based super learner would lead to additional improvements owing to its more robust treatment of outliers.

In this simulation, we fixed the sample size at 1000 and the percentage of observations with zero costs at around 20\%. We considered three scenarios for outlier proportions (low-3.5\%, medium-10\%, high-20\%) and analyzed 1000 simulated data sets for each scenario. Costs were simulated using a Tweedie distribution with varying values assigning to mean ($\mu$), power ($p$) and dispersion ($\sigma^2$) parameters to control for fixed zero-inflation (20\%) as well as different skewness (Appendix A.4 of the supplementary material). The true ATE was calculated numerically in each setting.

A TMLE of the ATE requires two inputs -- an estimate of the probability of the intervention $X_1$ as a function of $X_2, \dots, X_5$ (the so-called propensity score) and an estimate of the conditional mean of the costs as a function of $X_1, \dots, X_5$ (the so-called outcome regression). Each TMLE considered in our simulation used a main term logistic regression for the propensity score. We note that this is a correct model specification owing to the randomized treatment assignment. We considered three different approaches to modeling the outcome regression: a standard super learner (TMLE-standard SL), the partial-CV Huber loss-based super learner (TMLE-partial CV Huber SL), and the nested-CV Huber loss-based super learner (TMLE-nested CV Huber SL). Each super learner consisted of the same ten candidate algorithms and was based on ten-fold cross-validation (Appendix Table 7 of the supplementary material). 

For the Huber loss-based super learner we proposed a ``global'' set for $\lambda$ that included 37 candidate values equally spaced from 0.1 to 1$e$8. In each simulated data set, the set used for $\lambda$ selection was a subset of the ``global'' set with the upper bound determined by the maximum observed value of the outcome. 

We compared the performance of the unadjusted sample mean difference to the three super learner-based TMLEs in terms of bias, variance, and mean squared error of the point estimates. 

Table 2 summarized the results for ATE estimation using different estimators. Overall, the unadjusted sample mean difference has smaller bias but larger variance and larger MSE when compared with the three TMLEs, indicating the expected gain in precision from adjusting for prognostic covariates. Among the three TMLEs, the TMLE that used super learner with Huber loss and nested CV for tuning parameter selection had the smallest bias, variance and MSE, while the TMLE that used super learner with Huber loss and partial CV exhibited slightly higher bias and variance, but still lower than the TMLE based on the standard super learner. Breaking the results down further based on outlier proportion, we see that gains in performance were modest in settings with low skewness, but pronounced in settings with medium and high skewness. For example, in the high skewness setting, the MSE of the estimates based on a standard super learner was reduced by 11.4\% and 9.4\% by utilizing a super learner with Huber loss with nested/partial CV, respectively.

\section{Data analysis}

In this section we applied our proposed method on MEPS data and BOLD data. We used the same super learner library with four algorithms as in the simulations for analyzing the MEPS data while used a different super learner library with eight algorithms for analyzing the BOLD data. 

\subsection{MEPS Data}

MEPS is a national survey on the financing and use of medical care of families and individuals, their medical providers, and employers across the United States \citep{cohen2003design, cohen2009medical}. We used the longitudinal data of MEPS 2016-2017, with 2016 MEPS used as a training sample and the same individuals in 2017 MEPS used as a testing sample. We developed a prediction model for total annual healthcare expenditures based on 14 covariates including demographics, medical conditions, and insurance characteristics (Appendix Table 9 of the supplementary material). Final sample included adults with non-missing data and contained 10,925 observations for training sample and 10,815 observations for testing sample. Distributions of total expenditures in 2016 \& 2017 were both zero-inflated, with about 20\% (train: 17.83\%; test: 19.56\%) of observations having zero expenditures, as well as highly right skewed, with over 10\% (train: 11.80\%; test: 12.25\%) of observations having outlying expenditures (Appendix Table 8 and Appendix Figure 2 of the supplementary material). Two-stage estimators were also modeled on MEPS data with the same super learner library as the simulations. We set up 35 candidate values of $\lambda$ ranging from 0.1 to 350,000 for $\lambda$ selection and fix $V=10$ for all the $V$-fold cross-validation involved in building various estimators. 

In MEPS data, we again found that nested-CV Huber loss-based super learner is the best performing algorithm (Table 3), although its improvements in MSE over partial-CV Huber loss-based super learner (0.3\%) and standard super learner (1.9\%) are modest compared to simulations. Appendix Table 10 of the supplementary material presents the results of the two-stage estimators, which are similar to the one-stage estimators as Huber loss, as well as the nested-CV procedure provided extra benefits in terms of cost estimation.

\subsection{BOLD Data}

The BOLD project is a large, community-based registry of patients aged 65 years and older who presented with primary care visits for a new episode of back pain from March 2011 to March 2013 \citep{jarvik2012study}. Expenditures in BOLD data were calculated as total relative value units (RVUs), a measure of value used in the US Medicare reimbursement formula for physician services \citep{glass2002relative}. We used the total annual RVUs one year after index as the outcome in this study and the model was developed based on 24 covariates from baseline patient questionnaires and EHR (Appendix Table 11 of the supplementary material). Final sample contained 4,397 observations, with around 10\% of them being outliers (Appendix Table 12 and Appendix Figure 3 of the supplementary material). No zero total annual RVUs was observed in the BOLD data, necessitating the implementation of one-stage estimators with super learner library including different algorithms that are more appropriate to model the positive outcome (Appendix Table 13 of the supplementary material). We set up 25 candidate values of $\lambda$ ranging from 0.1 to 3000 for $\lambda$ selection and fix $V=10$ for all the V-fold cross-validation involved in building the one-stage estimators. To evaluate the model, the data were randomly partitioned into ten distinct folds. All the one-stage estimators were trained using nine of the folds (training fold) and predictions were obtained using the remaining fold (validation fold). The performance was estimated as the MSE of the predictions on the validation fold, averaged over the ten validation folds. Specifically, $\lambda$ selection by two cross-validation procedures was conducted on both training fold and validation fold to obtain the training $\lambda$ ($\lambda_t$) and the validation $\lambda$ ($\lambda_v$). Intraclass Correlation Coefficient (ICC) between $\lambda_t$ and $\lambda_v$ was calculated to evaluate the two procedures for $\lambda$ selection.

Prediction of future RVUs using only baseline covariates is challenging, as evidenced by the small $R^2$ values for the six learners, ranging from 0.0270 to 0.0551 (Table 4). Similar to MEPS data, we observed modest prediction improvements in the Huber loss-based learners over the standard learners (MSE improvements: 1.3\% - 1.9\% for discrete learner; 0.6\% - 1.1\% for super learner) and nested-CV over partial-CV (MSE improvements: 0.6\% for discrete learner; 0.5\% for super learner). In addition, nested-CV is more consistent and robust than partial-CV in terms of $\lambda$ selection, with higher ICC for both Huber loss-based discrete learner (nested-CV vs. partial-CV: 0.052 vs. 0.009) and Huber loss-based super learner (nested-CV vs. partial-CV: 0.860 vs. 0.339) (Appendix Table 14 of the supplementary material).

\section{Discussion}

Our results indicate that cross validation-based learning provides an optimal way of selecting estimators when the research goal is optimizing Huber risk for a given value of robustification parameter. However, we have also demonstrated empirically that the proposed procedure is useful even in settings where optimizing based on mean squared error is the ultimate goal. This latter setting has wider implications as it relates to super learning performed in the context of effect estimation and other semiparametric estimation problems. These results indicate that more robust finite sample behavior can be obtained by selecting the robustification tuning parameter using an additional layer of cross validation. However, we also demonstrate that a non-nested cross validation scheme can be employed that improves performance of ensemble learners \emph{at minimal computational cost}. Our simulation and data analyses indicate that the expected gains associated with our procedure over standard super learner practice are most pronounced in settings with limited sample size or a high fraction of outlying values. In settings with both these features, the gains associated with our method can be dramatic.

Our method has the potential to be extended to more general robust loss functions \citep{barron2019general}. Additionally, our approach used grid search for selection of robustification parameter in Huber loss. This may cause issues for settings where computational complexity associated with computing weights is high. An alternative approach is to develop a selection scheme that avoids the grid search while adapting the robustification parameter according to data.


\section{Software}

Software in the form of R code, together with a sample
input data set and complete documentation is available on
request from the corresponding author (ziyue.wu@emory.edu). Our methods described in this article are available in the R function HuberSL on the GitHub repository https://github.com/wuziyueemory/Huber-loss-based-SuperLearner.

{\it Conflict of Interest}: None declared.

\section*{Funding}

This research was supported in part under award 1R01HL137808 from the National Heart Lung and Blood Institute of the United States National Institute of Health. The findings and conclusions in this report are those of the authors and do not necessarily represent the views of of the National Institutes of Health.

\bibliographystyle{biom}
\bibliography{refs}

\clearpage

\begin{table}[h!]

\caption{Overall results of six one-stage learners in simulations
}
{\tabcolsep=4pt
\begin{tabular}{@{}cccccccccc@{}}
& \multicolumn{3}{c}{Discrete Learner} && \multicolumn{3}{c}{Super Learner}  \\ \cline{2-4} \cline{6-8}
Metric & Standard & Huber & Huber && Standard &
Huber & Huber \\
&& (partial-CV) & (nested-CV) &&& (partial-CV) & (nested-CV)\\
MSE ($10^{10}$) & 0.5566 & 0.5365 & 0.5216 && 0.5285 & 0.5168 & 0.5053 \\
Relative MSE (\%) && 96.39\% & 93.71\% &&& 97.79\% & 95.61\% \\
$R^2$ & 0.3824 & 0.4071 & 0.4306 && 0.4101 & 0.4290 & 0.4483 \\
RE (\%) && 106.46\% & 112.60\% &&& 104.61\% & 109.31\% \\
MAE & 42671.72 & 42080.86 & 41668.15 && 41634.26 & 41339.34 & 41043.41
\end{tabular}}
\end{table}

\begin{table}[h!]
\caption{Results of ATE estimation with different learners in Simulation
}
{\tabcolsep=6pt
\begin{tabular}{@{}cccccccccc@{}}
&& Unadjusted & & TMLE &TMLE \\
Skewness & Metrics & Mean & TMLE& Huber SL & Huber SL \\
&& Difference &Standard SL& (partial-CV) & (nested-CV)
\\
Overall & Bias & 72.20 & 222.44 & 124.26 & 118.67 \\
(11.27\%) & Variance ($10^8$) & 1.0098 & 0.8250 & 0.7916     & 0.7846\\
& MSE ($10^8$) & 1.0101 & 0.8255 & 0.7920 & 0.7849 \\
& Relative MSE (\%) & 122.36\% & 100.00\% & 95.95\% & 95.08\% \\
\hline

Low & Bias & -137.25 & -85.37 & -249.30 & -170.44 \\
(3.47\%) & Variance ($10^8$) & 1.4866 & 1.3758 & 1.3686 & 1.3660 \\
& MSE ($10^8$) & 1.4869 & 1.3767 & 1.3700 & 1.3671 \\
& Relative MSE (\%) & 108.00\% & 100.00\% & 99.52\% & 99.30\% \\
\hline

Medium & Bias & 81.14 & 304.95 & 263.49 & 215.93 \\
(10.80\%) & Variance ($10^7$) & 4.6932 & 2.5780 & 2.4330 & 2.4192 \\
& MSE ($10^7$) & 4.6947 & 2.5883 & 2.4408 & 2.4244 \\
& Relative MSE (\%) & 181.38\% & 100.00\% & 94.30\% & 93.67\% \\
\hline

High & Bias & 272.72 & 447.75 & 358.60 & 310.52 \\
(19.53\%) & Variance ($10^8$) & 1.0753 & 0.8430 & 0.7641 & 0.7475 \\
& MSE ($10^8$) & 1.0763 & 0.8451 & 0.7655 & 0.7486 \\
& Relative MSE (\%) & 127.35\% & 100.00\% & 90.58\% & 88.58\%

\end{tabular}}
{\parbox{\textwidth}{
\vspace{2em}
\footnotesize SL = super learner. Relative MSE = MSE / MSE of TMLE Standard SL $\times$ 100. Zero percentage = 19.40\%, 18.30\%, 19.85\% and 20.04\% and true value of ATE = 23,884, 23,652, 23,998, and 24,002 for overall, low, medium and high skewness, respectively.}
}
\end{table}

\begin{table}[h!]
\caption{Results of six one-stage learners in MEPS analysis
}
{\tabcolsep=4.1pt
\begin{tabular}{@{}cccccccccc@{}}
& \multicolumn{3}{c}{Discrete Learner} && \multicolumn{3}{c}{Super Learner}  \\ \cline{2-4} \cline{6-8}
Metric & Standard & Huber & Huber && Standard &
Huber & Huber \\
&& (partial-CV) & (nested-CV) &&& (partial-CV) & (nested-CV) \\
MSE ($10^{8}$) & 2.3358 & 2.2887 & 2.2728 && 2.2896 & 2.2525 & 2.2453 \\
Relative MSE (\%) && 97.98\% & 97.30\% &&& 98.38\% & 98.07\% \\
$R^2$ & 0.1046 & 0.1227 & 0.1287 && 0.1223 & 0.1365 & 0.1393 \\
RE (\%) && 117.30\% & 123.04\% &&& 111.61\% & 113.90\% \\
MAE & 5888.40 & 5837.57 & 5832.21 && 5835.65 & 5790.34 & 5795.78
\end{tabular}}
\end{table}

\begin{table}[h!]
\caption{Results of six one-stage learners in BOLD analysis
}
{\tabcolsep=4.1pt
\begin{tabular}{@{}cccccccccc@{}}
& \multicolumn{3}{c}{Discrete Learner} && \multicolumn{3}{c}{Super Learner}  \\ \cline{2-4} \cline{6-8}
Metric & Standard & Huber & Huber && Standard &
Huber & Huber \\
&& (partial-CV) & (nested-CV) &&& (partial-CV) & (nested-CV)\\
MSE & 17574.87 & 17341.13 & 17237.43 && 17253.12 & 17147.93 & 17066.29 \\
Relative MSE (\%) && 98.67\% & 98.08\% &&& 99.39\% & 98.92\% \\
RE (\%) && 147.78\% & 169.26\% &&& 112.95\% & 122.99\% \\
MAE & 49.93 & 49.68 & 49.61 && 49.57 & 49.46 & 49.50
\end{tabular}}
\end{table}

\begin{figure}[h!]
\centering
\includegraphics[width=1.\textwidth]{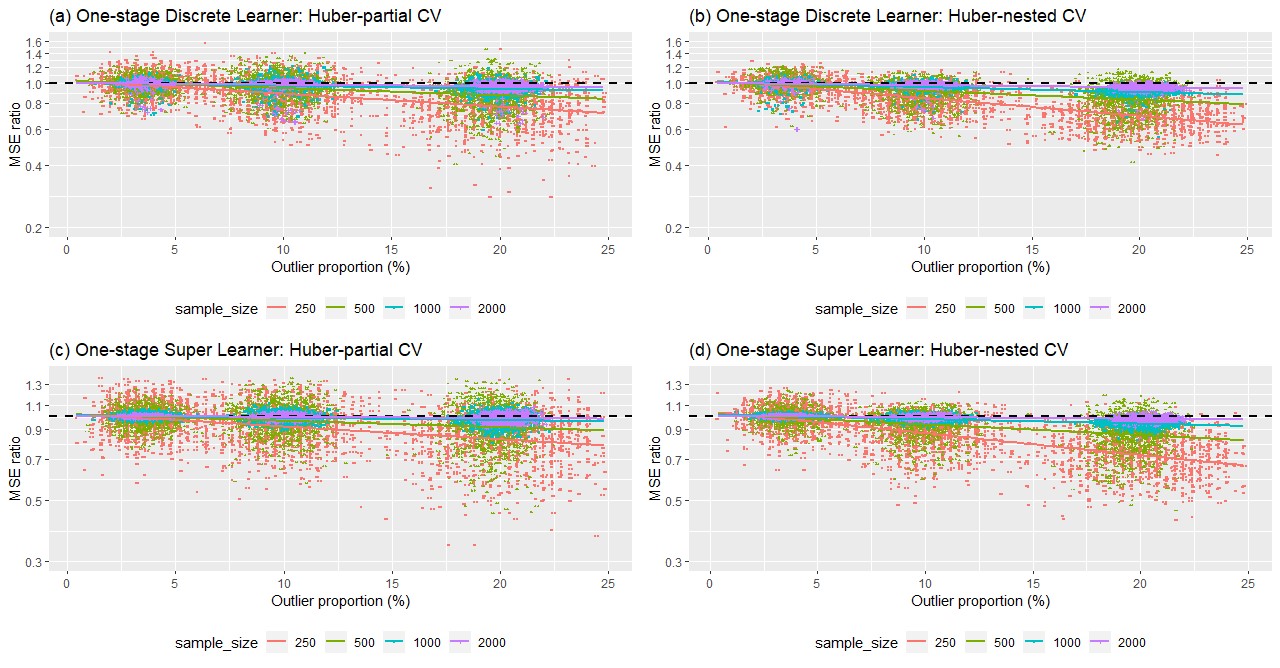}
\caption{Scatter plot of MSE ratio (log-scale on Y axis) between one-stage (a) partial-CV Huber Discrete Learner vs. standard discrete learner; (b) nested-CV Huber Discrete Learner vs. standard discrete learner; (c) partial-CV Huber Super Learner vs. standard super learner; (d) nested-CV Huber Super Learner vs. standard super learner; over different skewness and sample size in simulations. X-axis is the proportion of outliers in simulated data, Y-axis is the MSE ratio. Points with different colors in scatter plot refer to observed  MSE ratio under different sample size. Lines with different colors in scatter plot refer to the regression lines of corresponding points under different sample size. Black horizontal dashed line refers to a MSE ratio of 1 for reference.}
\end{figure}

\clearpage

\begin{center}
\LARGE \textbf{Supplementary Material}\\
\end{center}

\noindent\textbf{A.1. Theoretical Justification}

Our theoretical justification follows from \citet{van2007super} and \citet{van2006oracle}. We first introduce and revise some notations. Suppose one observes \textit{n} i.i.d. observations $O_{i}=(X_{i}, Y_{i})\sim P, i=1, \dots ,n$, where $Y$ is a scalar outcome and $X$ is a vector of covariates. Under the Huber loss function with a certain robustification parameter $\lambda \in R^+$, let us denote the \textit{true} super learner $Q^\text{SL}_{\alpha_0}$ as the true Huber risk minimizer by
$$Q^\text{SL}_{\alpha_0}=\mbox{arg}\mbox{min}_{Q^\text{SL}_{\alpha}} E\left\{L_{\lambda}(Q^\text{SL}_{\alpha})\right\} =  \mbox{arg}\mbox{min}_{Q^\text{SL}_{\alpha}} R_{\lambda}(Q^\text{SL}_{\alpha})$$
where $$L_{\lambda}(Q)(X,Y)=\left\{\begin{matrix}
\frac{1}{2}\left\{Y-Q(X)\right\}^2 & |Y-Q(X)| \leq \lambda  \\
\lambda(|Y-Q(X)| - \frac{1}{2}\lambda) & |Y-Q(X)| > \lambda
\end{matrix}\right.\ , \ \lambda \in R^+$$
and
$$
R_{\lambda}(Q) = \int L_{\lambda}(Q)(x,y)dP(x,y) \ .
$$

For super learner, suppose there is a collection of $K$ individual algorithms with the corresponding prediction functions $(\hat{Q}_1,\dots, \hat{Q}_K)$. The super learner is parameterized as $\hat{Q}_{\alpha}^\text{SL} = \sum_{i=1}^K \alpha_k \hat{Q}_k$, where $\alpha$ is the vector of coefficients (weights) that linearly combines the $K$ algorithms. Without loss of generality, we assume the weights $\alpha$ to be constrained as $\alpha_i \in [0,1], \ i=1,\dots,K$ and $\sum_{k=1}^K \alpha_i=1$. The parameter space of $\alpha$ can be written as $\mathcal{A} = [0,1]^K$. We consider a grid (or any finite subset) $\mathcal{A}_n$ of $\alpha$ values in the bounded parameter space $\mathcal{A}$, and let $K(n)=\#\mathcal{A}_n$ be the number of grid points such that $K(n)\leq n^q$ for some constant $q<\infty$. We then consider $\mathcal{Q} = \{\hat{Q}_{\alpha}^\text{SL}: \alpha \in \mathcal{A}_n\}$ as a collection of candidate super learners. 

$V$-fold cross-validation splits the data into a training sample and a validation sample. Denote $\hat{Q}_{k,v},\ k=1,\dots,K, \ v=1,\dots,V$ as the predict functions with candidate algorithm $k$ fitted on the $v$-th training sample and $\hat{Q}_{\alpha,v}^\text{SL}$ as the super learner estimated from the $v$-th training sample. 

The oracle weight vector $\tilde{\alpha}$ is the one that minimizes the true Huber risk on the unknown distribution $P$ given $V$ training-sample-specific fits:
$$
\tilde{\alpha}=\mbox{arg}\mbox{min}_{\alpha \in \mathcal{A}_n} \sum_{v=1}^V \int L_{\lambda}(\hat{Q}_{\alpha, v}^\text{SL})(o) dP(o) = \mbox{arg}\mbox{min}_{\alpha\in \mathcal{A}_n} \sum_{v=1}^V R_{\lambda}(\hat{Q}_{\alpha, v}^\text{SL}) \ .
$$ 
The oracle weights represent the best weights that could be selected at a population-level, given the training samples seen. These weights can be used to construct the oracle super learner $\hat{Q}^\text{SL}_{\tilde{\alpha}}$.

In Huber loss-based super learner (also known as cross-validated estimator), the cross-validation replaces the unknown true distribution $P$ by the validation sample distribution $\hat{P}_V$ and calculates $\hat{\alpha}$ as the Huber loss-based super learner weights that minimize the cross-validated estimate of the Huber risks, defined as:
$$
\hat{\alpha}=\mbox{arg}\mbox{min}_{\alpha\in \mathcal{A}_n} \sum_{v=1}^V \int L_{\lambda}(\hat{Q}_{\alpha, v}^\text{SL})(o) d\hat{P}_V(o) = \mbox{arg}\mbox{min}_{\alpha\in \mathcal{A}_n} \sum_{v=1}^V \hat{R}_{\lambda}(\hat{Q}_{\alpha, v}^\text{SL}) \ .
$$
where
$$\hat{R}_{\lambda}(\hat{Q}_{\alpha, v}^\text{SL}) = \frac{1}{|\mathcal{I}_v|} \sum_{i \in \mathcal{I}_v} L_{\lambda}(\hat{Q}_{\alpha, v}^{\text{SL}})(O_i) \ .
$$

The goal of this section is to theoretically prove that for a given $\lambda \in R^+$, the Huber loss-based super learner $\hat{Q}^{\text{SL}}_{\hat{\alpha}}$ asymptotically converges to the oracle super learner $\hat{Q}^{\text{SL}}_{\tilde{\alpha}}$ in terms of the Huber risk averaged over $V$ training-sample-specific fits. \citet{van2006oracle} (Theorem 2.3) established the following inequality for the risk averaged over the $V$-fold cross-validation data splits between the cross-validated estimator ($\hat{Q}^{\text{SL}}_{\hat{\alpha}}$) and the oracle estimator ($\hat{Q}^{\text{SL}}_{\tilde{\alpha}}$).

\textbf{Theorem 1}
(\citet{van2006oracle}) For $\hat{Q}^\text{SL}_{\alpha} \in \mathcal{Q}$. Let $\left\{M(\hat{Q}^\text{SL}_{\alpha}),v(\hat{Q}^\text{SL}_{\alpha})\right\}$ be a Bernstein pair for the function $o\rightarrow L(\hat{Q}^\text{SL}_{\alpha})(o)$ and assume that $R(\hat{Q}^\text{SL}_{\alpha})=\int L(\hat{Q}^\text{SL}_{\alpha})(o)dP(o) \geq 0$ for every $\hat{Q}^\text{SL}_{\alpha} \in \mathcal{Q}$. Then for any $\delta>0$ and $1 \leq p \leq 2$,
  \begin{multline*}
    \sum_{v=1}^V R(\hat{Q}^\text{SL}_{\hat{\alpha},v}) \leq (1+2\delta)\left\{ \sum_{v=1}^V R(\hat{Q}^\text{SL}_{\tilde{\alpha},v})\right\} +
    (1+\delta)E\left\{ \frac{16}{(n^1)^{\frac{1}{p}}}\right\} \\
    \times \log(1+\#\mathcal{Q}) \sup_{\hat{Q}^\text{SL}_{\alpha} \in \mathcal{Q}}\left[ \frac{M(\hat{Q}^\text{SL}_{\alpha})}{(n^1)^{1-1/p}}+\left\{\frac{v(\hat{Q}^\text{SL}_{\alpha})}{R(\hat{Q}^\text{SL}_{\alpha})^{2-p}}\right\}^{1/p}\left(\frac{1+\delta}{\delta}\right)^{2/p-1} \right]
  \end{multline*}
where $n^1$ is the size of the validation sample. The condition that risk $R(\hat{Q}^\text{SL}_{\alpha})\geq0$ is met under the Huber loss function since it's a combination of squared error loss and absolute error loss, both of which are non-negative. Recall that for a measurable function $f:\mathcal{O} \rightarrow \mathcal{R}$, $\left\{M(f),v(f)\right\}$ is a pair of Bernstein numbers if:
$$
M(f)^2P\left\{e^{\frac{|f|}{M(f)}}-1-\frac{|f|}{M(f)} \right\} \leq \frac{1}{2}v(f)
$$
And it was shown in \citet{van2006oracle} that if $f$ is uniformly bounded, then $(||f||_{\infty},1.5Pf^2)$ is a pair of Bernstein numbers.

\textbf{Assumption A1}
Assume there exists a constant $C_0<\infty$, such that both individual algorithms $\hat{Q}_k,\ k=1,\dots,K$ and outcome $Y$ are bounded: $|Y|\leq C_0$ for all $Y \in \mathcal{Y}$ and $\mbox{sup}_{k}\mbox{sup}_{x \in \mathcal{X}}|\hat{Q}_k(x)| \leq C_0$.

\textbf{Lemma 1}
For Huber loss function given a particular $\lambda$: $o \rightarrow L_{\lambda}(\hat{Q}^\text{SL}_{\alpha})(o)$, where $\hat{Q}^\text{SL}_{\alpha}: x \rightarrow \mathcal{R}$ and $\lambda \in R^+$, its Bernstein pairs $\left\{M(\hat{Q}^\text{SL}_{\alpha}),v(\hat{Q}^\text{SL}_{\alpha})\right\}$ under assumptions A1 satisfy:
$$M(\hat{Q}^\text{SL}_{\alpha}) = \mbox{max}\left(2C_0^2, \ 2\lambda C_0 \right)$$
and
$$v(\hat{Q}^\text{SL}_{\alpha}) = 1.5 \times E\left\{ L_{\lambda}^2 (\hat{Q}^\text{SL}_{\alpha})\right\}$$
Furthermore,
$$v(\hat{Q}^\text{SL}_{\alpha}) \leq 1.5 \times M(\hat{Q}^\text{SL}_{\alpha}) \times R_{\lambda}(\hat{Q}^\text{SL}_{\alpha})$$

\noindent\textbf{Proof.} The Huber loss function can be rewritten as:
\begin{align*}
& L_{\lambda}(\hat{Q}^\text{SL}_{\alpha})(o)= \\
&\hspace{0.2in} \mathbbm{1}\left\{|Y-\hat{Q}^\text{SL}_{\alpha}(X)| \leq \lambda \right\}\frac{1}{2}\left\{Y-\hat{Q}^\text{SL}_{\alpha}(X)\right\}^2+\mathbbm{1}\left\{| Y-\hat{Q}^\text{SL}_{\alpha}(X)| > \lambda\right\}\lambda\left\{| Y-\hat{Q}^\text{SL}_{\alpha}(X)| - \frac{1}{2}\lambda\right\}
\end{align*}
Under (A1), $\sup_{\hat{Q}^\text{SL}_{\alpha} \in \mathcal{Q}}\sup_{X \in \mathcal{X}}|\hat{Q}^\text{SL}_{\alpha}(X)| \leq C$ for some $C < \infty$. We have $\left\{Y-\hat{Q}^\text{SL}_{\alpha}(X)\right\}^2 \leq 4C_0^2$ and $|Y-\hat{Q}^\text{SL}_{\alpha}(X)| \leq 2C_0$. Therefore,
$$
L_{\lambda}(\hat{Q}^\text{SL}_{\alpha})(o) \leq \mathbbm{1}\left\{|Y-\hat{Q}^\text{SL}_{\alpha}(X)| \leq \lambda\right\}2C_0^2 + \mathbbm{1}\left\{|Y-\hat{Q}^\text{SL}_{\alpha}(X)| > \lambda\right\}2\lambda C_0
$$
As a result, the Huber loss function $L_{\lambda}(\hat{Q}^\text{SL}_{\alpha})(o)$ is bounded by $\mbox{max}\left(2C_0^2, \ 2\lambda C_0\right)$. And it's easy to show that $R_{\lambda}(\hat{Q}^\text{SL}_{\alpha}) \leq \mbox{max}\left(2C_0^2, \ 2\lambda C_0\right)$ as $R_{\lambda}(\hat{Q}^\text{SL}_{\alpha}) = E\left\{ L_{\lambda}(\hat{Q}^\text{SL}_{\alpha}) \right\}$. The corresponding Bernstein pairs for the Huber loss function $L_{\lambda}\left\{(Y,\hat{Q}^\text{SL}_{\alpha}(X)\right\}$ is:
$$M(\hat{Q}^\text{SL}_{\alpha})=\mbox{max}\left(2C_0^2, \ 2\lambda C_0 \right)$$
and
$$v(\hat{Q}^\text{SL}_{\alpha})=1.5 \times E\left\{ L_{\lambda}^2(\hat{Q}^\text{SL}_{\alpha})\right\}$$

The second moment of the Huber loss function has the following form and upper bound:
  \begin{align*}
    E\left\{ L_{\lambda}^2(\hat{Q}^\text{SL}_{\alpha}) \right\}
    & = E\left[ \mathbbm{1}\left\{|Y-\hat{Q}^\text{SL}_{\alpha}(X)| \leq \lambda\right\}\left\{\frac{1}{2}\left(Y-\hat{Q}^\text{SL}_{\alpha}(X)\right)^2 \right\}^2 \right] + \\
    & \hspace{0.8in} E\left[ \mathbbm{1}\left\{| Y-\hat{Q}^\text{SL}_{\alpha}(X)| > \lambda\right\}\left\{\lambda \left(| Y-\hat{Q}^\text{SL}_{\alpha}(X)| - \frac{1}{2}\lambda\right) \right\}^2 \right] \\
    & \leq E\left[ \mathbbm{1}\left\{|Y-\hat{Q}^\text{SL}_{\alpha}(X)| \leq \lambda \right\}\frac{1}{2}\left\{Y-\hat{Q}^\text{SL}_{\alpha}(X)\right\}^2 \times 2C_0^2\right] + \\
    & \hspace{0.8in} E\left[\mathbbm{1}\left\{| Y-\hat{Q}^\text{SL}_{\alpha}(X)| > \lambda\right\}\lambda\left\{| Y-\hat{Q}^\text{SL}_{\alpha}(X)| - \frac{1}{2}\lambda\right\} \times 2\lambda C_0 \right] \\
    & \leq \mbox{max}\left(2C_0^2, \ 2\lambda C_0 \right) \times R_{\lambda}(\hat{Q}^\text{SL}_{\alpha}) \\
    & =M(\hat{Q}^\text{SL}_{\alpha}) \times R_{\lambda}(\hat{Q}^\text{SL}_{\alpha})
  \end{align*}

Therefore,

$$v(L_{\lambda}(\hat{Q}^\text{SL}_{\alpha})=1.5 \times E\left\{ L_{\lambda}^2(\hat{Q}^\text{SL}_{\alpha})\right\} \leq 1.5 \times M(\hat{Q}^\text{SL}_{\alpha})\times R(\hat{Q}^\text{SL}_{\alpha})$$

We apply Theorem 1 \citep{van2006oracle} with Lemma 1, so that for the Huber Loss function, we have

\textbf{Theorem 2}
Let $\hat{\alpha}$ be the super learner weights of the Huber loss-based super learner $\hat{Q}^\text{SL}_{\hat{\alpha}}$ and $\tilde{\alpha}$ be the oracle weights for the oracle super learner $\hat{Q}^\text{SL}_{\tilde{\alpha}}$. For a collection of super learners in $\mathcal{Q} = \{\hat{Q}_{\alpha}^\text{SL}: \alpha \in \mathcal{A}_n\}$ where $\hat{Q}^\text{SL}_{\alpha} = \sum_{k=1}^K \alpha_k\hat{Q}_k$ and the Huber loss $L_{\lambda}(Q)$ for a given $\lambda \in R^+$, there is 
$$
\sum_{v=1}^V R_{\lambda}(\hat{Q}_{\hat{\alpha}, v}^\text{SL}) \leq \sum_{v=1}^V R_{\lambda}(\hat{Q}_{\tilde{\alpha}, v}^\text{SL}) + O\left(\frac{\log(n)}{n^{1/2}}\right) \ .
$$
where $R_{\lambda}(\hat{Q}_{\hat{\alpha}, v}^\text{SL})$ and $R_{\lambda}(\hat{Q}_{\tilde{\alpha}, v}^\text{SL})$ are the Huber risk averaged over V training-sample-specific fits for the Huber loss-based super learner and the oracle super learner respectively.

\textbf{Proof.}
Write $C=max\left(2C_0^2, \ 2 \lambda C_0 \right)$. Recall we assumed the cardinality of $\mathcal{A}_n$ as $K(n) \leq n^q$, and by definition, $\#\mathcal{Q} = \#\mathcal{A}_n$. Hence, we have $\log(1+\#\mathcal{Q}) \leq q\log(n)$. Therefore, applying the inequality in Lemma 1 to theorem 1, for any $\delta>0$ and $1 \leq p \leq 2$, there is
  \begin{multline*}
    \sum_{v=1}^V R_{\lambda}(\hat{Q}_{\hat{\alpha}, v}^\text{SL}) \leq (1+2\delta)\left\{\sum_{v=1}^VR_{\lambda}(\hat{Q}_{\tilde{\alpha}, v}^\text{SL})\right\} +
    (1+\delta)E\left\{ \frac{16}{(n^1)^{1/p}}\right\} \\
    \times
    \log(1+\#\mathcal{Q})
     \sup_{\hat{Q}^\text{SL}_{\alpha} \in \mathcal{Q}}\left[ \frac{C}{(n^1)^{1-1/p}}+\left\{1.5C\times R_{\lambda}(\hat{Q}^\text{SL}_{\alpha})^{p-1}\right\}^{1/p}\left(\frac{1+\delta}{\delta}\right)^{2/p-1} \right]
  \end{multline*}

Furthermore, by $R_{\lambda}(\hat{Q}^\text{SL}_{\alpha}) \leq C$ and $\#\mathcal{Q} \leq n^q$, we have
\begin{align*}
& \sum_{v=1}^V R_{\lambda}(\hat{Q}_{\hat{\alpha}, v}^\text{SL}) \leq (1 + 2\delta) \left\{\sum_{v=1}^V R_{\lambda}(\hat{Q}_{\tilde{\alpha}, v}^\text{SL})\right\} + \\
& \hspace{1.0in} (1+\delta)E\left\{ \frac{16}{(n^1)^{1/p}}\right\} \times q\log(n) \times \left\{ \frac{C}{(n^1)^{1-1/p}}+(1.5C^p)^{1/p}\left(\frac{1+\delta}{\delta}\right)^{2/p-1} \right\} \\
& \hspace{0.8in} \leq (1 + 2\delta) \left\{\sum_{v=1}^VR_{\lambda}(\hat{Q}_{\tilde{\alpha}, v}^\text{SL})\right\} + \\
& \hspace{1.0in} (1+\delta)E\left\{ \frac{16}{(n^1)^{1/p}}\right\}
    \times q\log(n) \times \left\{ \frac{C}{(n^1)^{1-1/p}}+1.5C\Big(\frac{1+\delta}{\delta}\Big)^{2/p-1} \right\}
\end{align*}

Sample size of the validation sample is approximately a fixed positive proportion of the entire data, so $n^1=O(n)$. Let $p=2$ and $\delta=\frac{1}{n^{1/2}}$, the above inequality becomes
\begin{align*}
& \hspace{-0.6in} \sum_{v=1}^V R_{\lambda}(\hat{Q}_{\hat{\alpha}, v}^\text{SL}) \leq \sum_{v=1}^VR_{\lambda}(\hat{Q}_{\tilde{\alpha}, v}^\text{SL}) + \frac{2}{n^{1/2}}VC \ + \\
& \hspace{0.8in} (1+\frac{1}{n^{1/2}})E\left\{ \frac{16}{(n^1)^{1/2}}\right\} \times q\log(n) \times \left\{\frac{C}{(n^1)^{1-1/2}}+1.5C \right\} \\
& \hspace{0.2in} = \sum_{v=1}^VR_{\lambda}(\hat{Q}_{\tilde{\alpha}, v}^\text{SL}) + O\Big(\frac{\log(n)}{n^{1/2}}\Big)
\end{align*}

$\frac{\log(n)}{n^{1/2}}$ asymptotically goes to zero. As long as
\begin{equation}
\frac{\frac{\log(n)}{n^{1/2}}}{\sum_{v=1}^VR_{\lambda}(\hat{Q}_{\tilde{\alpha}, v}^\text{SL})} \rightarrow 0 \quad as \quad n \rightarrow \infty
\end{equation}

Then Huber loss-based super learner $\hat{Q}_{\hat{\alpha}, v}^\text{SL}$ is asymptotically equivalent to the oracle super learner $\hat{Q}_{\tilde{\alpha}, v}^\text{SL}$ in terms of the Huber risks averaged over the $V$-fold cross-validation data splittings when fitting the estimators on the training sample, that is
$$\frac{\sum_{v=1}^V R_{\lambda}(\hat{Q}_{\hat{\alpha}, v}^\text{SL})}{\sum_{v=1}^V R_{\lambda}(\hat{Q}_{\tilde{\alpha}, v}^\text{SL})} \rightarrow 1 \quad as \quad n \rightarrow \infty$$
When equation (9.3) does not hold, then $\hat{Q}_{\hat{\alpha}, v}^\text{SL}$ achieves the $\frac{\log(n)}{n^{1/2}}$ rate,
$$\sum_{v=1}^V R_{\lambda}(\hat{Q}_{\hat{\alpha}, v}^\text{SL}) = O\Big(\frac{\log(n)}{n^{1/2}}\Big)$$

Given assumption A1 and Lemma 1, the Huber loss function satisfies the two assumptions of theorem 3.1 proposed by \citet{van2007super} regarding the asymptotic optimality of the standard super learner. Let $p=1$, $n^1=np$ where $p$ denotes the proportion of validation sample over the entire data set and define a constant $C(\delta)=16(1+\delta)\left\{ C+1.5C(\frac{1+\delta}{\delta}) \right\}$ we have for any $\delta>0$
\begin{equation*}
  \begin{array}{l}
    \sum_{v=1}^VR_{\lambda}(\hat{Q}_{\hat{\alpha}, v}^\text{SL}) \leq (1+2\delta)\left\{\sum_{v=1}^VR_{\lambda}(\hat{Q}_{\tilde{\alpha}, v}^\text{SL})\right\}+C(\delta)\frac{\log \left(1+K(n)\right)}{np}
  \end{array}
\end{equation*}
which is similar to the finite sample result of Theorem 1 from \citet{van2007super}.
\break

\noindent\textbf{A.2. Regression formula for cost estimation in simulation study}

Costs were simulated using a two-stage procedure to allow for zero-inflation. In stage-1, we drew a random variable $Z$ from a \textit{Bernoulli} distribution with the probability of zero determined by a logistic model with parameters $\mu_{\beta}(X)$:
$$
\log\Big(\frac{Pr(Z=0|X)}{1-Pr(Z=0|X)}\Big) = \mu_{\beta}(X) \ , \ X=(X_1, \dots, X_5)
$$
where $\mu_\beta(X)$ was fixed at $\mu_\beta(X) = 0.6 + 0.1(X_1 + X_2 - X_3 + X_4 - X_5 + X_1 X_2 - X_2 X_3 + X_3 X_4 - X_4 X_5)$ in our study to yield a zero percentage of around 35\%. If $Z = 0$, then we set costs equal to zero. If $Z=1$, we first drew the cost from a gamma distribution with $Y \sim \Gamma(shape = 10 |\mu_k(X)|, scale = 1.5)$ where $\mu_k(X) = X_1 + X_2 + X_3 + X_4 + X_5 + X_1 X_2 + X_2 X_3 + X_3 X_4 + X_4 X_5$. Then we keep the cost below the upper quartile of cost data ($Y[Y \leq Y_{Q_3}]$) as is and make certain changes to the cost beyond the upper quartile of cost data ($Y[Y > Y_{Q_3}]$) depending on specific outlier proportion and sample size (Table 5). An scaling procedure is involved to ensure that all the simulated cost data are within the same scale ($1\sim1,000,000$).

\begin{table}[h!]
\caption{Positive cost distribution under different outlier proportion \& sample size}
{\tabcolsep=4pt
\begin{tabular}{@{}cccccccccc@{}}
Outlier proportion & Sample size & Positive cost distribution \\
Low & All & $Y$ \\ \hline
& All & $Y[Y \leq Y_{Q_3}] = Y$ \\
Medium & 250 & $Y[Y > Y_{Q_3}] \sim Y + \Gamma(shape = 1.13 \ \mu^2_k(X), scale = 1.5)$ \\
& 500/1000/2000 & $Y[Y > Y_{Q_3}] \sim Y + \Gamma(shape = 0.71 \ \mu^2_k(X), scale = 1.5)$ \\ \hline
& All & $Y[Y \leq Y_{Q_3}] = Y$ \\
High & 250 & $Y[Y > Y_{Q_3}] \sim Y + \Gamma(shape = 38 \  \mu^2_k(X), scale = 1.5)$ \\
& 500/1000/2000 & $Y[Y > Y_{Q_3}] \sim Y + \Gamma(shape = 2.9 \  \mu^2_k(X), scale = 1.5)$
\end{tabular}}
{\parbox{5.2in}{
\footnotesize Note: $Y$ follows a Gamma distribution with $Y \sim \Gamma(shape = 10  |\mu_k(X)|, scale = 1.5)$ where $\mu_k(X) = X_1 + X_2 + X_3 + X_4 + X_5 + X_1 X_2 + X_2 X_3 + X_3  X_4 + X_4 X_5$.}
}
\end{table}
\break

\noindent\textbf{A.3. Individual algorithms for two-stage super learner}

The two-stage learners (standard discrete learner, standard super learner, partial-CV Huber loss-based discrete learner, partial-CV Huber loss-based super learner, nested-CV Huber loss-based discrete learner, nested-CV Huber loss-based super learner) were included in the simulation study and MEPS analysis to compare their performance to the corresponding one-stage learners. For two-stage learners, we consider four algorithms at stage-1, four algorithms at stage-2, and four single-stage algorithms, resulting in a total of twenty algorithms. The algorithms include a mixture of machine learning algorithms such as random forests and parametric regressions such as GLM (Table 6). For each two-stage estimator, we use the same candidate values of robustification parameter $\lambda$ and set $V=10$ for all the $V$-fold cross-validation involved.

\begin{table}[h!]
\centering
\caption{Individual algorithms for two-stage super learner in simulation}
{\tabcolsep=20pt
\begin{tabular}{lcl} 
Super Learner & Stage & Individual Algorithm\\
& 1 & GLM (logistic regression) \\
& 1 & Lasso (logistic regression) \\
& 1 & K-Nearest Neighbor (KNN) \\
Two-stage & 1 & Random Forest \\ 
\cline{2-3}
& 2 & GLM (log link, Gamma family) \\
& 2 & Log OLS + smearing \\
& 2 & Xgboost \\
& 2 & Random forest \\ 
\hline
& & OLS \\
One-stage & & Lasso (OLS) \\
& & Support Vector Machine (SVM) \\
& & Random Forest
\end{tabular}}
\end{table}
\break

\noindent\textbf{A.4. Regression formula for causal effect estimation in simulation study}

For causal effect estimation, outcomes were simulated using a Tweedie distribution. Tweedie distribution is a family of probability distributions which includes the Normal, Gamma, Poisson and Inverse Gaussian distributions. It is defined as subfamily of exponential dispersion models (ED) with a special mean-variance relationship. Though probability density function cannot be evaluated directly, a random variable $Y$ is Tweedie distributed $Tw_p(\mu, \sigma^2)$ if $Y \sim ED(\mu, \sigma^2)$ with mean $\mu = E(Y)$, positive dispersion parameter $\sigma^2$ and variance $Var(Y) = \sigma^2 \mu^p$, where $p \in \mathcal{R}$ is called the Tweedie power parameter. In this simulation, mean ($\mu$) was determined by a parameter $\mu_{\alpha}(X)$ with $\mu_{\alpha}(X) = X_1 + X_2 + X_3 + X_1 X_4 + X_1 X_5 + X_2 X_3 + X_4 X_5$, power ($p$) was restricted as $1<p<2$ (Compound Poisson distribution) and dispersion ($\sigma^2$) was assigned with different constants to control for fixed zero-inflation (20\%) as well as different outlier proportions (Table 7). Additionally, outcomes generated under different outlier proportions were multiplied by a different constant $C$ to keep the true ATE approximately the same in each scenario.

\begin{table}[h!]
\caption{Outcome distribution under different outlier proportion in simulation}
\begin{center}
\begin{tabular}{@{}cccccccccc@{}}
Sample size & Outlier proportion & Outcome distribution \\
& Low & $Y \sim 9200 \ Tw_{p=1.5}(\mu=15+|\mu_{\alpha}(X)|, \sigma^2 = 5)$ \\
1000 & Medium & $Y \sim 1000 \ Tw_{p=1.5}(\mu=\mu_{\alpha}^2(X), \sigma^2 = 1.9)$ \\ 
& High & $Y \sim 1000 \ Tw_{p=1.932}(\mu=\mu_{\alpha}^2(X), \sigma^2 = 10)$
\end{tabular}
\end{center}
{\parbox{4.5in}{
\footnotesize Note: $\mu_{\alpha}(X) = X_1 + X_2 + X_3 + X_1X_4 + X_1X_5 + X_2X_3 + X_4X_5$}
}
\end{table}

\noindent\textbf{A.5. Details of covariates used in the BOLD study}

In this study, patient self-reported questionnaire responses were used to predict future expenditures (measured by RVUs) in several categories. The covariates include the following measures from patient self-reported questionnaires collected at baseline: (1) Socio-demographics (age, sex, race, ethnicity, education, employment status, etc.); (2) Pain-related characteristics (back/leg pain duration, back/leg pain intensity, modified Roland-Morris Disability Questionnaire, Brief Pain Inventory Activity Interference Scale); (3) PHQ-4 measure of anxiety and depressive symptoms; (4) European Quality of Life 5 Dimension (EQ5D) index and Visual Analog Scale; (5) Number of falls; and (6) Recovery expectation. Besides, we also include the Quan comorbidity score, baseline diagnosis, and total RVUs at one year before index visit from EHR as covariates. 

Most patients in the BOLD study were females (64.9\%), Caucasians (73.4\%), recruited from Kaiser site (66.6\%), high school graduate (54.6\%), non-smokers (55.1\%), retired (81.8\%), and lived with spouse or partner (59.5\%) with a mean age of 74 years old. With respect to back pain related measures, patients in the BOLD study were more likely to be diagnosed with back pain only (67.6\%), have no falls (92.5\%) and no injuries (96.6\%) in the last 3 weeks, and have had back pain for less than 3 months (53.8\%). They rated their back pain (5.0) and leg pain (3.5) as moderate intensity, they reported minor pain interference with activities (BPI: 3.4), slight psychological distress (PHQ-4: 1.6), a modest level of disability (RMDQ: 9.7), medium back pain recovery expectation (5.5) and good quality of life, as measured by the EQ-5D index (0.76) and EQ-5D VAS (74.4). The mean total RVUs in a year before index visit is 39.1. To reduce the sparsity in data, excess levels in certain baseline categorical covariates were grouped together to facilitate the modeling process. This grouping procedure was consistent with previous BOLD studies. For (1) race: collapse Native American Indian or Native Alaskan, Native Hawaiian or other Pacific Islander, Other and Multiple raced checked as ‘Mixed race’; (2) Education: collapse High school graduate or obtained a GED, Vocational or technical or trade school, Some college as ‘$\geq$ High school’; (3) Marital: collapse Married, Living with a partner as ‘Living with spouse or partner’ and Separated, Divorced, Never married and presently single, Widowed as ‘Not living with spouse or partner’; (4) Employment: collapse Working full-time, Working part-time as ‘Working Full-time/Part-time’ and On leave of absence, Unemployed and looking for work, Homemaker, Other as ‘Other’; (5) Falls in last 3 weeks: collapse all positive numbers of falls in last 3 weeks as ‘Patient with one or more falls in last three weeks’ (6) Quan comorbidity score: collapse Quan comorbidity score from 2 to 17 as ‘2 and more’.

\newpage

\section*{\textbf{Appendix - Tables}}

\begin{table}[h!]
\caption{Results of six one-stage learners stratified by sample size in simulations
}
{\tabcolsep=4pt
\begin{tabular}{@{}cc|ccc|cccccc@{}}
&&& Discrete Learner &&& Super Learner && \\
Sample & Metric & Standard & Huber & Huber & Standard &
Huber & Huber \\
size &&& (partial-CV) & (nested-CV) && (partial-CV) & (nested-CV)\\
& MSE ($10^{10}$) & 0.7929 & 0.7375 & 0.6972 & 0.7353 & 0.7022 & 0.6714 \\
& Relative MSE (\%) && 93.01\% & 87.93\% && 95.50\% & 91.30\% \\
250 & $R^2$ & 0.2351 & 0.2811 & 0.3383 & 0.2732 & 0.3179 & 0.3662 \\
& RE (\%) && 119.55\% & 143.88\% && 116.38\% & 134.07\% \\
& MAE & 48934.31 & 47348.54 & 46677.12 & 46700.84 & 46263.58 & 45852.12 \\
\hline

& MSE ($10^{10}$) & 0.6074 & 0.5916 & 0.5773 & 0.5728 & 0.5637 & 0.5523 \\
& Relative MSE (\%) && 97.41\% & 95.06\% && 98.41\% & 96.41\% \\
500 & $R^2$ & 0.3652 & 0.3946 & 0.4189 & 0.4064 & 0.4241 & 0.4431 \\
& RE (\%) && 108.05\% & 114.69\% && 104.35\% & 109.04\% \\
& MAE & 43534.32 & 43312.36 & 43124.70 & 42343.18 & 42333.14 & 42228.25 \\
\hline

& MSE ($10^{10}$) & 0.4549 & 0.4487 & 0.4448 & 0.4389 & 0.4358 & 0.4325 \\
& Relative MSE (\%) && 98.64\% & 97.79\% && 99.28\% & 98.53\% \\
1000 & $R^2$ & 0.4349 & 0.4495 & 0.4588 & 0.4578 & 0.4664 & 0.4746 \\
& RE (\%) && 103.36\% & 105.50\% && 101.88\% & 103.68\% \\ 
& MAE & 39913.76 & 39750.08 & 39682.68 & 39242.05 & 39202.23 & 39154.17 \\
\hline

& MSE ($10^{10}$) & 0.3712 & 0.3682 & 0.3670 & 0.3669 & 0.3654 & 0.3650 \\
& Relative MSE (\%) && 99.18\% & 98.86\% && 99.61\% & 99.49\% \\
2000 & $R^2$ & 0.4944 & 0.5031 & 0.5065 & 0.5031 & 0.5076 & 0.5091 \\
& RE (\%) && 101.75\% & 102.43\% && 100.91\% & 101.21\% \\
& MAE & 38272.42 & 38003.11 & 38000.03 & 37883.01 & 37867.36 & 37897.80
\end{tabular}}
{\parbox{5.3in}{
\footnotesize Note: All metrics are calculated based on mean value. }
}
\end{table}

\begin{table}[h!]
\caption{Results of six one-stage learners stratified by skewness in simulations
}
{\tabcolsep=4pt
\begin{tabular}{@{}cc|ccc|cccccc@{}}
&&& Discrete Learner &&& Super Learner && \\
Skewness & Metric & Standard & Huber & Huber & Standard &
Huber & Huber \\
&&& (partial-CV) & (nested-CV) && (partial-CV) & (nested-CV)\\
& MSE ($10^{10}$) & 0.9708 & 0.9652 & 0.9583 & 0.9355 & 0.9341 & 0.9302 \\
& Relative MSE (\%) && 99.42\% & 98.71\% && 99.85\% & 99.44\% \\
Low & $R^2$ & 0.4960 & 0.4990 & 0.5023 & 0.5136 & 0.5145 & 0.5163 \\
& RE (\%) && 100.61\% & 101.28\% && 100.19\% & 100.54\% \\
& MAE & 50343.41 & 50306.40 & 50233.93 & 49473.14 & 49486.45 & 49439.43 \\
\hline

& MSE ($10^{10}$) & 0.4258 & 0.4031 & 0.3853 & 0.3986 & 0.3848 & 0.3720 \\
& Relative MSE (\%) && 94.69\% & 90.49\% && 96.54\% & 93.33\% \\
Medium & $R^2$ & 0.3894 & 0.4185 & 0.4411 & 0.4248 & 0.4424 & 0.4585 \\
& RE (\%) && 107.47\% & 113.30\% && 104.16\% & 107.94\% \\
& MAE & 39663.08 & 39016.39 & 38545.65 & 38435.65 & 38159.88 & 37908.79 \\
\hline

& MSE ($10^{10}$) & 0.2732 & 0.2411 & 0.2213 & 0.2513 & 0.2315 & 0.2136 \\
& Relative MSE (\%) && 88.25\% & 81.00\% && 92.10\% & 84.97\% \\
High & $R^2$ & 0.2619 & 0.3038 & 0.3483 & 0.2920 & 0.3300 & 0.3700 \\
& RE (\%) && 116.00\% & 133.02\% && 113.04\% & 126.72\% \\
& MAE & 33730.16 & 32975.34 & 31076.84 & 32410.13 & 31456.70 & 30554.06
\end{tabular}}
{\parbox{5.3in}{
\footnotesize Note: All metrics are calculated based on mean value. }
}
\end{table}

\begin{table}[h!]
\caption{Overall results of 12 one-stage \& two-stage learners in Simulations
}
{\tabcolsep=4pt
\begin{tabular}{@{}cc|ccc|cccccc@{}}
&&& Discrete Learner &&& Super Learner && \\
Model & Metric & Standard & Huber & Huber & Standard &
Huber & Huber \\
&&& (partial-CV) & (nested-CV) && (partial-CV) & (nested-CV)\\
& MSE ($10^{10}$) & 0.5566 & 0.5365 & 0.5216 & 0.5285 & 0.5168 & 0.5053 \\
One & Relative MSE (\%) && 96.39\% & 93.71\% && 97.79\% & 95.61\% \\
Stage & $R^2$ & 0.3824 & 0.4071 & 0.4306 & 0.4101 & 0.4290 & 0.4483 \\
& RE (\%) && 106.46\% & 112.60\% && 104.61\% & 109.31\% \\ 
& MAE & 42671.72 & 42080.86 & 41668.15 & 41634.26 & 41339.34 & 41043.41 \\ \hline
& MSE ($10^{10}$) & 0.5445 & 0.5140 & 0.4921 & 0.5124 & 0.4966 & 0.4833 \\
Two & Relative MSE (\%) && 94.40\% & 90.38\% && 96.92\% & 94.32\% \\
Stage & $R^2$ & 0.3957 & 0.4366 & 0.4719 & 0.4283 & 0.4558 & 0.4800 \\
& RE (\%) && 110.34\% & 119.26\% && 106.42\% & 112.07\% \\
& MAE & 42445.09 & 41442.13 & 40738.02 & 41236.34 & 40769.19 & 40387.78
\end{tabular}}
{\parbox{5.3in}{
\footnotesize Note: All metrics are calculated based on mean value. }
}
\end{table}

\begin{table}[htbp]
\centering
\caption{Individual algorithms in one-stage super learner library for causal effect estimation in simulation}
{\tabcolsep=28pt
\begin{tabular}{cc} 
\hline
Super Learner & Algorithms\\
\hline
              & OLS \\
              & Lasso (OLS) \\
              & Multivariate Adaptive Regression Splines \\
              & Zero-Inflated Poisson (ZIP) \\
One-stage     & Tobit \\
              & Tweedie \\
              & Support Vector Machine \\
              & Random Forest \\
              & Boosted regression trees (Xgboost) \\
              & Neural Network\\
              \hline
\end{tabular}}
\end{table}

\begin{table}[htbp]
\centering
\caption{Summary statistics for annual healthcare expenditures of MEPS data}
{\tabcolsep=22pt
\begin{tabular}{lcc} 
\hline
Measures & 2016 MEPS (N=10925) & 2017 MEPS (N=10815)\\
\hline
0.05 Percentile & 0 & 0 \\
0.25 Percentile & 157 & 130 \\
0.50 Percentile & 1076 & 1156 \\
0.75 Percentile & 4521 & 4902 \\
0.95 Percentile & 24277.8 & 28189.3 \\
Mean & 5368.4 & 6099.7 \\
SD & 13815.1 & 15992.3 \\
Skewness & 7.2 & 6.7 \\
Zero percentage (\%) & 17.8 & 19.6 \\
Outlier proportion (\%) & 11.80 & 12.25\\
\hline
\end{tabular}}
{\parbox{5.5in}{
\footnotesize The distribution of annual healthcare expenditures is highly skewed with a large mass at zero. The skewness is 7.2 and 6.7 for 2016 \& 2017 MEPS data, respectively (compare to 0 for symmetric data). Almost 20\% of observations have zero expenditures and in a very small fraction of observations, 2\% to be precise, had expenditures over \$50,000.}
}
\end{table}

\begin{table}[htbp]
\centering
\caption{Descriptive statistics of baseline covariates in MEPS data}
{\tabcolsep=17pt
\begin{tabular}{llcc} 

Covariate & & 2016 MEPS & 2017 MEPS  \\
 & & N=10925 & N=10815\\
Age & & 46.6 & 46.4 \\
Sex & Female (\%) & 45.6 & 45.5 \\
    & Male (\%) & 54.4 & 54.5 \\
Race & Hispanic (\%) & 28.8 & 28.9 \\
     & White (\%) & 42.5 & 42.4 \\
     & Black (\%) & 18.4 & 18.3 \\
     & Asian (\%) & 7.5 & 7.5 \\
     & Other (\%) & 2.9 & 2.9 \\
Region & Northeast (\%) & 16.3 & 16.2 \\
       & Midwest (\%) & 19.4 & 19.3 \\
       & South (\%) & 37.9 & 38 \\
       & West (\%) & 26.4 & 26.5 \\
Education (yrs) & & 12.9 & 12.9 \\
Income as percent poverty (\%) & & 361.1 & 369.5 \\
Private insurance (\%) & & 59.5 & 61 \\
Medicare (\%) & & 22.6 & 23.5 \\
Public insurance (\%) & & 22.9 & 23.6 \\
Uninsured (\%) & & 12.3 & 11 \\
Diabetes (\%) & & 12 & 11.9 \\
Hypertension (\%) & & 34.1 & 33.9 \\
Cancer (\%) & & 9 & 8.7 \\
Heart disease (\%) & & 13.1 & 12.8
\end{tabular}}
{\parbox{5.5in}{
\footnotesize
A smaller proportion of females (45.6\% for 2016 and 45.5\% for 2017) appeared in the MEPS analysis and the mean age was about 46.5 years. The 2016 MEPS data had similar education levels, race distributions, and region distributions as the 2017 MEPS data. Participants in 2017 MEPS had a better insurance and medical care coverage than participants in 2016 MEPS — likely because they are richer with a higher average family income as \% of the poverty line (369.5\% vs. 361.1\%). Participants of 2017 MEPS are also healthier compare to participants of 2016 MEPS with a slightly fewer presence of diabetes (11.9\% vs. 12.0\%), hypertension (33.9\% vs. 34.1\%), cancer (8.7\% vs. 9.0\%), and heart disease (12.8\% vs. 13.1\%).
}
}
\end{table}

\begin{table}[h!]
\caption{Results of 12 one-stage \& two-stage learners in MEPS analysis
}
{\tabcolsep=4.1pt
\begin{tabular}{@{}cc|ccc|cccccc@{}}
&&& Discrete Learner &&& Super Learner && \\
Model & Metric & Standard & Huber & Huber & Standard &
Huber & Huber \\
&&& (partial-CV) & (nested-CV) && (partial-CV) & (nested-CV)\\
& MSE ($10^{9}$) & 2.3358 & 2.2887 & 2.2728 & 2.2896 & 2.2525 & 2.2453 \\
One & Relative MSE (\%) && 97.98\% & 97.30\% && 98.38\% & 98.07\% \\
Stage & $R^2$ & 0.1046 & 0.1227 & 0.1287 & 0.1223 & 0.1365 & 0.1393 \\
& RE (\%) && 117.30\% & 123.04\% && 111.61\% & 113.90\% \\
& MAE & 5888.40 & 5837.57 & 5822.21 & 5835.65 & 5795.78 & 5790.34 \\ \hline

& MSE ($10^{9}$) & 2.2737 & 2.2129 & 2.1924 & 2.2357 & 2.1938 & 2.1840 \\
Two & Relative MSE (\%) && 97.33\% & 96.42\% && 98.13\% & 97.69\% \\
Stage & $R^2$ & 0.1284 & 0.1517 & 0.1596 & 0.1430 & 0.1591 & 0.1628 \\
& RE (\%) && 118.15\% & 124.30\% && 111.26\% & 113.85\% \\ 
& MAE & 5817.36 & 5749.64 & 5728.46 & 5773.34 & 5727.23 & 5718.58 
\end{tabular}}
\end{table}

\begin{table}[htbp]
\centering
\caption{Descriptive statistics of baseline covariates in BOLD data}
{\tabcolsep=18pt

\begin{tabular}{llc} 
Covariate & & No. (\%) of Patients  \\
& & N=4397\\
\hline
Study Site & Harvard Vanguard & 682 (15.5\%) \\
           & Henry Ford & 787 (17.9\%) \\
           & Kaiser & 2928 (66.6\%) \\
Age, mean (SD) & & 73.7 (6.8) \\
Gender - Female & & 2852 (64.9\%) \\
Hispanic - Yes & & 259 (5.9\%) \\
Race & Black & 671 (15.3\%) \\
     & Asian & 185 (4.2\%) \\
     & White & 3229 (73.4\%) \\
     & Mixed race & 312 (7.1\%) \\
Education &  High school & 252 (5.7\%) \\
          & = High school & 2399 (54.6\%) \\
          & College graduate & 972 (22.1\%) \\
          & Graduate degree & 774 (17.6\%) \\
\multicolumn{2}{l}{Living with spouse or partner} & 2616 (59.5\%) \\
Smoking status & Never Smoked & 2424 (55.1\%) \\
 & Quit  1 year ago & 1712 (38.9\%) \\
 & Current smoker/quit  1 year ago & 261 (5.9\%) \\
Employment & Working Full-time/Part-time & 483 (11.0\%) \\
           & Retired (not due to ill health) & 3598 (81.8\%) \\
           & Retired/disabled due to ill health & 125 (2.8\%) \\
           & Other & 191 (4.3\%) \\
Lawyer - Yes & & 26 (0.6\%) \\
Back Pain duration &  1 month & 1488 (33.8\%) \\
                   & 1 - 3 months & 879 (20.0\%) \\
                   & 3 - 6 months & 296 (6.7\%) \\
                   & 6 - 12 months & 257 (5.8\%) \\
                   & 1 - 5 years & 645 (14.7\%) \\
                   &  5 years & 832 (18.9\%) \\
\multicolumn{2}{l}{Back Pain intensity (0-10), mean (SD)} & 5.0 (2.8) \\
\multicolumn{2}{l}{Leg Pain intensity (0-10), mean (SD)} & 3.5 (3.3) \\
\multicolumn{2}{l}{Back Pain Recovery Expectations in 3 months (0–10), mean (SD)} & 5.5 (3.7) \\
\multicolumn{2}{l}{Patients with one or more fall in the last 3 weeks} & 328 (7.5\%) \\
\multicolumn{2}{l}{Patients with an injury\textsuperscript{*} caused by falls} & 150 (3.4\%) \\
\multicolumn{2}{l}{RMDQ score (0-24), mean (SD)} & 9.7 (6.3) \\
\multicolumn{2}{l}{BPI interference (0-10), mean (SD)} & 3.4 (2.5) \\
\multicolumn{2}{l}{EQ-5D index (0-1), mean (SD)} & 0.76 (0.17) \\
\multicolumn{2}{l}{EQ-5D VAS (0-100), mean (SD)} & 74.4 (18.3) \\
\multicolumn{2}{l}{PHQ-4 score (0-12), mean (SD)} & 1.6 (2.5) \\
Baseline diagnosis & Back pain only & 2972 (67.6\%) \\
                   & Back and leg pain & 954 (21.7\%) \\
                   & Spinal Stenosis & 217 (4.9\%) \\
                   & Other & 254 (5.8\%) \\
Quan comorbidity score & 0 & 527 (12.0\%) \\
                       & 1 & 1726 (39.3\%) \\
                       & 2 and more & 2144 (48.8\%) \\
\multicolumn{2}{l}{RVUs in a year before the index, mean (SD)} & 39.1 (89.1)\\
\hline
\end{tabular}}
{\parbox{5.7in}{
\footnotesize
\textsuperscript{*} The injury was defined as limiting regular activities for at least a day or requiring a visit to a doctor.
}
}
\end{table}

\begin{table}[htbp]
\centering
\caption{Summary statistics for total annual RVUs one year after index of BOLD data}
{\tabcolsep=28pt
\begin{tabular}{lc} 
\hline
Measures & Total annual RVUs one year after index \\
\hline
0.05 Percentile & 4.41 \\
0.25 Percentile & 13.80 \\
0.50 Percentile & 28.21 \\
0.75 Percentile & 58.35 \\
0.95 Percentile & 188.11 \\
Mean & 62.21 \\
SD & 134.40 \\
Skewness & 7.70 \\
Zero percentage(\%) & 0 \\
Outlier proportion (\%) & 10.12\\
\hline
\end{tabular}}
{\parbox{5in}{
\footnotesize
Distributions of the annual RVUs one year after index are highly skewed with heavy upper tails. There are about 10\% of RVUs being outliers and no zero RVUs in the outcome.}
}
\end{table}

\bigskip

\begin{table}[htbp]
\centering
\caption{Individual algorithms for one-stage super learner in MEPS data}
{\tabcolsep=28pt
\begin{tabular}{cc} 
\hline
Super Learner & Algorithms\\
\hline
              & OLS                                \\
              & Lasso (OLS)                        \\
              & Log OLS + smearing                 \\
One-stage     & GLM (Log link, Gamma family)       \\
              & GLM (Identity link, Gamma family)  \\
              & Tweedie                            \\
              & Support Vector Machine             \\
              & Random Forest\\
              \hline
\end{tabular}}
\end{table}

\bigskip

\begin{table}[htbp]
\caption{Comparison of $\lambda$ obtained from train vs. holdout sample in BOLD analysis}
{\tabcolsep=9pt
\begin{tabular}{@{}cccccccccc@{}}
\hline
Metric & Grid-Search & Huber loss-based discrete learner & Huber loss-based super learner\\
\hline
Accuracy & partial-CV & 20\% & 10\% \\
& nested-CV & 30\% & 20\% \\ \hline
ICC & partial-CV & 0.0087 & 0.339 \\
& nested-CV & 0.0518 & 0.860\\
\hline
\end{tabular}}
\end{table}

\newpage
\section*{\textbf{Appendix - Figures}}

\begin{figure}[htbp]
    \centering
    \includegraphics[width=5.8in]{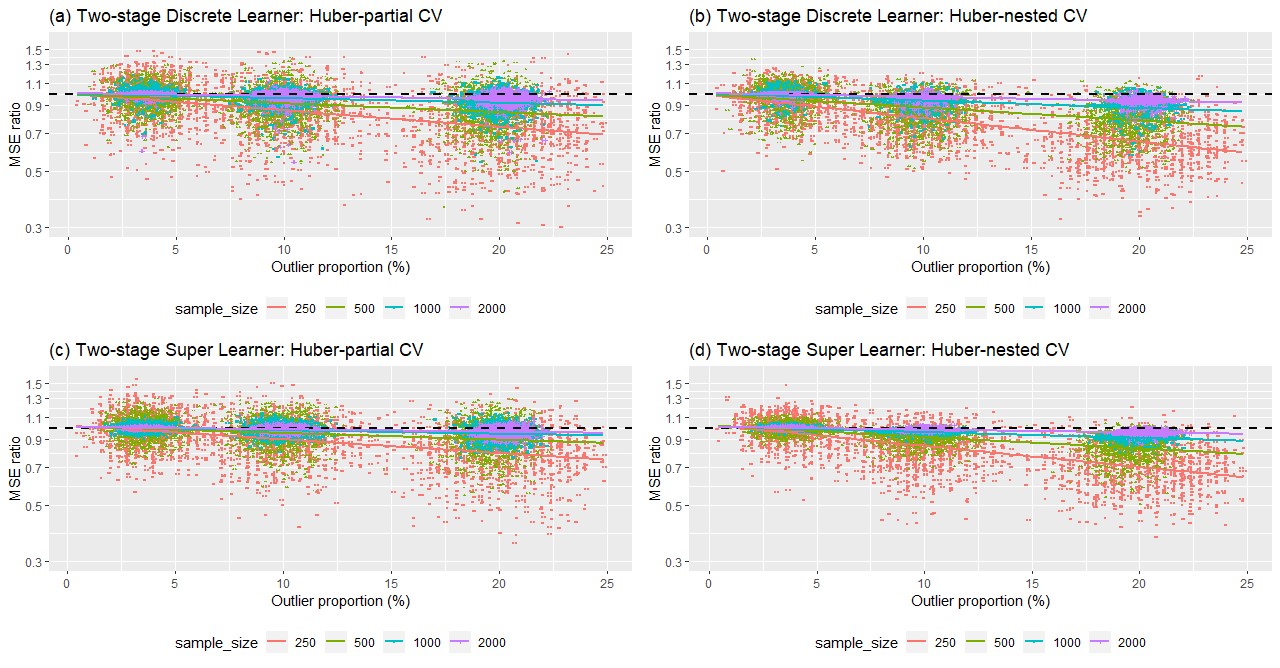}
    \caption{
    Scatter plot of MSE ratio (log-scale on Y axis) between two-stage (a) Huber Discrete Learner-partial CV vs. standard discrete learner; (b) Huber Discrete Learner-nested CV  vs. standard discrete learner; (c) Huber Super Learner-partial CV  vs. standard super learner; (d) Huber Super Learner-nested CV vs. standard super learner; over different outlier proportions and sample size in simulations. X-axis is the proportion of outliers in simulated data, Y-axis is the MSE ratio. Points with different colors in scatter plot refer to observed  MSE ratio under different sample size. Lines with different colors in scatter plot refer to the regression lines of corresponding points under different sample size. Black horizontal dashed line refers to a MSE ratio of 1 for reference. In all cases (Discrete Learner \& Super Learner; partial-CV \& nested CV): 1) the prediction improved as sample size increased; 2) the predictions degraded as outlier proportion increased; 3) the two huber loss-based learners (partial CV \& nested CV) is better than the standard squared loss-based estimators; 4) the nested-CV method is better than the partial-CV method. }
\end{figure}

\begin{figure}[htbp]
    \centering
    \includegraphics[width=5.5in]{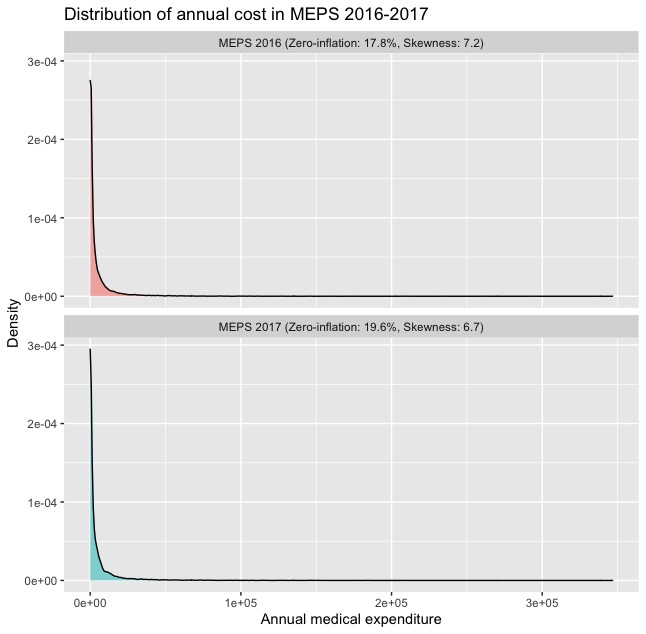}
    \caption{Distributions of outcomes in MEPS analysis. This figure illustrates the distributions of outcomes in MEPS analysis. All the outcomes in the MEPS analyses are highly skewed with a point mass at zero and heavy upper tails.}
\end{figure}

\begin{figure}[htbp]
    \centering
    \includegraphics[width=6in]{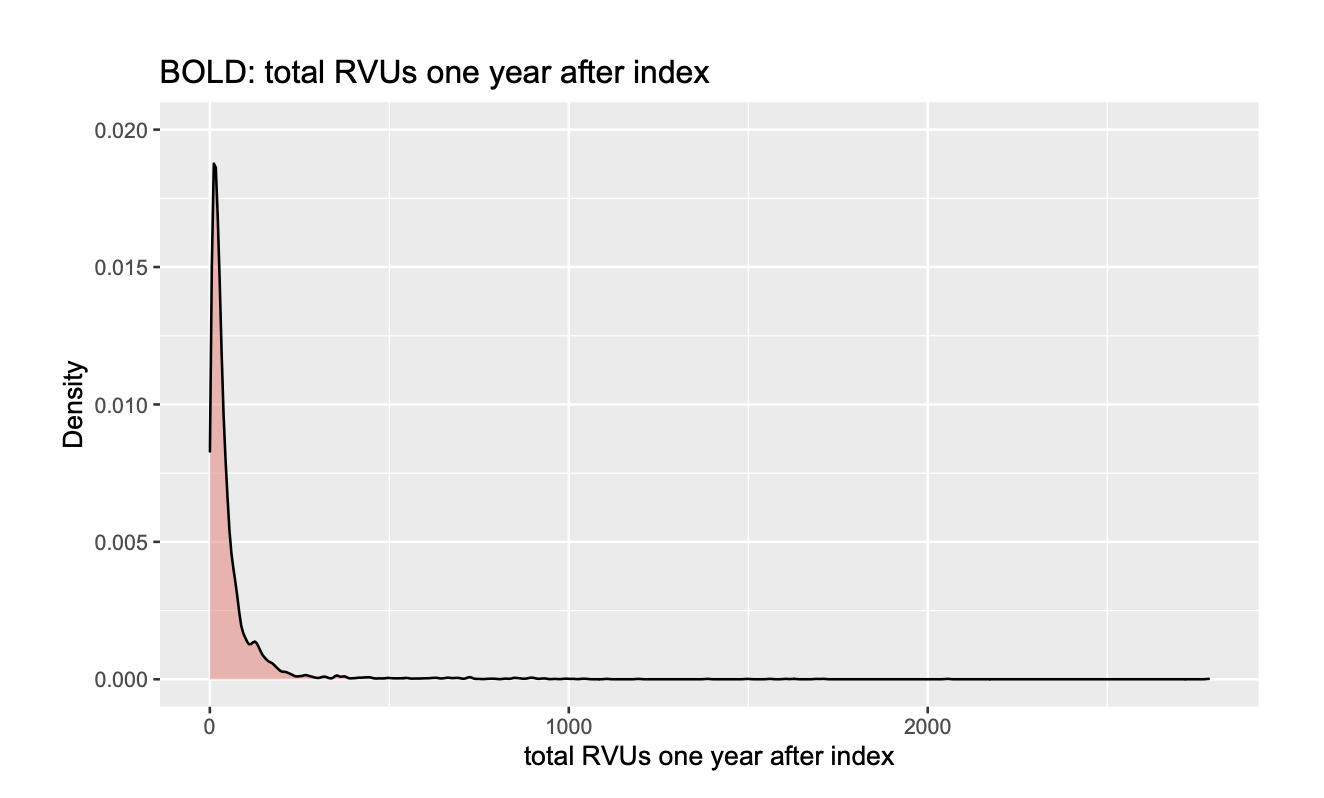}
    \caption{Distributions of annual total RVUs one year after index in BOLD analysis. The outcomes is highly skewed with heavy upper tails and outliers but no zero observation at all.}
\end{figure}

\FloatBarrier
\bibliographystyle{biorefs}
\bibliography{refs}

\end{document}